\documentclass{article}

\usepackage{arxiv}

\usepackage[utf8]{inputenc} 
\usepackage[T1]{fontenc}    
\usepackage{hyperref}       
\usepackage{url}            
\usepackage{booktabs}       
\usepackage{amsfonts}       
\usepackage{nicefrac}       
\usepackage{microtype}      
\usepackage{xcolor}         
\usepackage{graphicx}
\usepackage{subfigure}

\usepackage[ruled]{algorithm2e}
\usepackage{algorithmic}

\newtheorem{theorem}{Theorem}

\newtheorem{lemma}{Lemma}

\usepackage{bbm}        
\usepackage{amsmath}
\usepackage{comment}
\newcommand{\beq}{\begin{equation}}
\newcommand{\eeq}{\end{equation}}
\newcommand{\bea}{\begin{eqnarray}}
\newcommand{\eea}{\end{eqnarray}}
\newcommand{\bef}{\begin{figure}}
	\newcommand{\eef}{\end{figure}}
\newcommand{\bsc}{\begin{scriptstyle}}
	\newcommand{\esc}{\end{scriptstyle}}
\newcommand{\bd}{\begin{displaymath}}
\newcommand{\ed}{\end{displaymath}}

\DeclareMathOperator*{\argmax}{arg\,max}
\DeclareMathOperator*{\argmin}{arg\,min}

\title{Efficient Action Robust Reinforcement Learning with Probabilistic Policy Execution Uncertainty}

\author{%
    Guanlin Liu\\
    University of California, Davis\\
    One Shields Avenue, Davis, CA 95616\\
  \texttt{glnliu@ucdavis.edu} \\
  \And
    Zhihan Zhou\\
    Northwestern University\\
    633 Clark Street, Evanston, IL 60208\\
  \texttt{zhihanzhou@u.northwestern.edu} \\
\And
    Han Liu\\
    Northwestern University\\
    633 Clark Street, Evanston, IL 60208\\
  \texttt{hanliu@northwestern.edu} \\
    \And
  Lifeng Lai\\
  University of California, Davis\\
  One Shields Avenue, Davis, CA 95616\\
  \texttt{lflai@ucdavis.edu} \\
}

\begin{document}

\maketitle

\begin{abstract}
Robust reinforcement learning (RL) aims to find a policy that optimizes the worst-case performance in the face of uncertainties. In this paper, we focus on action robust RL with the probabilistic policy execution uncertainty, in which, instead of always carrying out the action specified by the policy, the agent will take the action specified by the policy with probability $1-\rho$ and an alternative adversarial action with probability $\rho$. We  establish the existence of an optimal policy on the action robust MDPs with probabilistic policy execution uncertainty and provide the action robust Bellman optimality equation for its solution. Furthermore, we develop Action Robust Reinforcement Learning with Certificates (ARRLC) algorithm that achieves minimax optimal regret and sample complexity. Furthermore, we conduct numerical experiments to validate our approach's robustness, demonstrating that ARRLC outperforms non-robust RL algorithms and converges faster than the robust TD algorithm in the presence of action perturbations.
\end{abstract}

\section{Introduction}

Reinforcement learning (RL), a framework of control-theoretic problem that makes decisions over time under an unknown environment, has many applications in a variety of scenarios such as recommendation systems \cite{RL_Recommendations}, autonomous driving \cite{RL_Car}, finance \cite{finrl2020} and business management \cite{RL_business}, to name a few.  However, the solutions to standard RL methods are not inherently robust to uncertainties, perturbations, or structural changes in the environment, which are frequently observed in real-world settings. A trustworthy reinforcement learning algorithm should be competent in solving challenging real-world problems with robustness against perturbations and uncertainties. 

Robust RL aims to improve the worst-case performance of algorithms deterministically or statistically in the face of uncertainties. The uncertainties could happen in different MDP components, including observations/states \cite{zhang2020robust,sun2022who}, actions \cite{tessler2019action,Klima2029robust}, transitions \cite{nilim2005robust,wang2021online}, and rewards \cite{huang2019deceptive,lecarpentier2019non}. Robust RL against action uncertainties focuses on the discrepancy between the actions generated by the RL agent and the conducted actions \cite{tessler2019action}. Thus, action uncertainties can be called as policy execution uncertainties. Taking the robot control as an example, such policy execution uncertainty may come from the actuator noise, limited power range, or actuator failures in the real world. \textcolor{black}{Taking the medication advice in healthcare as another example, such policy execution uncertainty may come from the patient's personal behaviors like drug refusal, forgotten medication, or overdose etc.}

Adversarial training~\cite{goodfellow2014explaining, madry2018towards} has been recognized as one of the most effective approaches in traditional supervised learning tasks in training time defenses. A lot of robust RL methods adopt the adversarial training framework and thus assume an adversary conducting adversarial attacks to mimic the naturalistic uncertainties~\cite{pinto2017robust, tessler2019action, Klima2029robust}. Training with an adversary can naturally be formulated as a zero-sum game between the adversary and the RL agent~\cite{tessler2019action}.

\cite{tessler2019action} proposed probabilistic action robust MDP (PR-MDP) in which, with probability $\rho$, an alternative adversarial action is taken. \cite{tessler2019action} introduced the Probabilistic Robust Policy Iteration (PR-PI) algorithm to train an adversary along with the agent. PR-PI algorithm converges toward the optimal value but requires a MDP solver to solve the optimal adversarial policy when the agent policy is given and the optimal agent policy when the adversarial policy is given. Thus, it is not suit for unknown reward functions or unknown transition probabilities. A similar idea as the PR-MDP was presented \cite{Klima2029robust}, which extends temporal difference (TD) learning algorithms by a new robust operator and shows that the new algorithms converge to the optimal robust $Q$-function. However, no theoretical guarantee on sample complexity or regret is given. 

In this paper, we aim to fill in the gaps of the existing work on policy execution uncertainties. In particular, we develop a minimax optimal sample efficient algorithm for action robust RL with probabilistic policy execution uncertainty. Our major contributions are summarized as follows:

\begin{itemize}
    \item We model the episodic RL with probabilistic policy execution uncertain set. We provide the action robust Bellman equation and the action robust Bellman optimality equation. We show that there always exists an optimal robust policy which is deterministic and can be solved via the induction of the action robust Bellman optimality equation. 
    \item We develop a new algorithm, Action Robust Reinforcement Learning with Certificates (ARRLC), for episodic action robust MDPs, and show that it achieves minimax order optimal regret and minimax order optimal sample complexity.
    \item We develop a model-free algorithm for episodic action robust MDPs, and analyze its  regret and sample complexity. 
    \item We conduct numerical experiments to validate the robustness of our approach. In our experiments, our robust algorithm achieves a much higher reward than the non-robust RL algorithm when being tested with some action perturbations; and our ARRLC algorithm converges much faster than the robust TD algorithm in \cite{Klima2029robust}.
\end{itemize}

\section{Related work}
We mostly focus on papers that are related to sample complexity bounds for the episodic RL and the two-player zero-sum Markov game, and action robust RL, that are close related to our model.
We remark that there are also related settings, e.g., infinite-horizon discounted MDP \cite{li2020sample, he2021nearly}, robust RL with other uncertainties \cite{lecarpentier2019non,zhang2020robust,wang2021online}, robust offline RL \cite{guo2022model,shi2022distributionally}, adversarial training with a generative RL model \cite{xu2023improved,panaganti2022sample}, adversarial attacks on RL \cite{Adaptive-Reward-Poisoning, liu2021provably, ICLR2021}, etc. These settings are beyond the scope of this paper, though our techniques may be also related to these settings.

\paragraph{Action robust RL} \cite{pinto2017robust} introduce robust adversarial reinforcement learning to address the generalization issues in reinforcement learning by training with a destabilizing adversary that applies disturbance forces to the system.  \cite{tessler2019action} introduce two new criteria of robustness for reinforcement learning in the face of action uncertainty. One is probabilistic action robust MDP (PR-MDP) in which, instead of the action specified by the policy, an alternative adversarial action is taken with probability $\rho$. Another is noisy action robust MDP (NR-MDP) criterion, in which a perturbation is added to the continues action vector itself. They generalize their policy iteration approach to deep reinforcement learning (DRL) and provide extensive experiments.
\cite{Klima2029robust} extends TD learning algorithms by a new robust operator and shows that the new algorithms converge to the optimal robust $Q$-function.

\paragraph{Sample Complexity Bounds for the Episodic RL}There is a rich literature on sample complexity guarantees for episodic tabular RL, for example~\cite{kearns2002near, strehl2006pac,auer2008near, UCBVI,dann2017unifying, jin2018q,dann2019policy,simchowitz2019non, zhang2020almost,zhang2021reinforcement}. Most relevant to our paper is the work about policy certificates~\cite{dann2019policy}.The algorithm outputs policy certificates that bound the sub-optimality and return of the policy in the next episode. They show that computing certificates can even improve the sample-efficiency of optimism-based exploration. 

\paragraph{Sample Complexity Bounds for the Two-player Zero-sum Markov Game} Training with an adversary can naturally be formulated as a zero-sum game between the adversary and the RL agent. Some sample efficient algorithms for two-player zero-sum Markov game can be used to train the action robust RL agent. \cite{liu2021sharp} design an algorithm named optimistic Nash value iteration (Nash-VI) for two-player zero-sum Markov games that is able to output an $\epsilon$-approximate Nash policy in $\widetilde{\mathcal{O}}(SABH^3/\delta^2)$ episodes of game playing. \cite{jin2021v} design a new class of fully decentralized algorithms V-learning, which provably learns $\epsilon$-approximate Nash equilibrium in $\widetilde{\mathcal{O}}(SAH^5/\delta^2)$ episodes of two-player zero-sum game playing. The two multi-agent RL algorithms can be used to solve the action robust optimal policy but are not minimax optimal. They are a factor of $A$ or $H^2$ above the minimax lower bound.

\section{Problem formulation} \label{sec:model}

\paragraph{Tabular MDPs} We consider a tabular episodic MDP $\mathcal{M} = (\mathcal{S}, \mathcal{A}, H, P, R)$, where $\mathcal{S}$ is the state space with $|\mathcal{S}| = S$, $\mathcal{A}$ is the action space with $|\mathcal{A}| = A$, $H \in \mathbb{Z}^+$ is the number of steps in each episode, $P$ is the transition matrix so that $P_h(\cdot|s,a)$ represents the probability distribution over states if action $a$ is taken for state $s$ at step $h \in [H]$, and $R_h:\mathcal{S} \times \mathcal{A} \rightarrow [0,1]$ represents the reward function at the step $h$. In this paper, the probability transition functions and the reward functions can be different at different steps. 

The agent interacts with the MDP in episodes indexed by $k$. 
Each episode $k$ is a trajectory $\{ s_{1}^k, a_{1}^k, r_{1}^k,  \cdots, s_{H}^k, a_{H}^k, r_{H}^k\}$ of $H$ states $s_{h}^{k} \in \mathcal{S}$, action $a_{h}^{k} \in \mathcal{A}$, and reward $r_{h}^{k} \in [0,1]$.
At each step $h \in [H]$ of episode $k$, the agent observes the state $s_{h}^{k}$ and chooses an action $a_{h}^{k}$. After receiving the action, the environment generates a random reward $r_{h}^{k} \in [0,1]$ derived from a distribution with mean $R_h(s_{h}^{k}, a_{h}^{k})$ and next state $s_{h+1}^{k}$ which is drawn from the distribution $P_h( \cdot | s_{h}^{k}, a_{h}^{k})$.
For notational simplicity, we assume that the initial states $s_{1}^k = s_1$ is deterministic in different episode $k$. 

A (stochastic) Markov policy of the agent is a set of $H$ maps $\pi := \{ \pi_{h}: \mathcal{S} \rightarrow \Delta_{\mathcal{A}} \}_{h \in [H]} $, where $\Delta_{\mathcal{A}}$ denotes the simplex over $\mathcal{A}$.
We use notation $\pi_h(a|s)$ to denote the probability of taking action $a$ in state $s$ under stochastic policy $\pi$ at step $h$. A deterministic policy is a policy that maps each state to a particular action. Therefore, when it is clear from the context, we abuse the notation $\pi_h(s)$ for a deterministic policy $\pi$ to denote the action $a$ which satisfies $\pi_h(a|s) = 1$. 

\paragraph{Action Robust MDPs} In the action robust case, the policy execution is not accurate and lies in some uncertainty set centered on the agent's policy $\pi$. Denote the actual behavior policy by $\widetilde{\pi}$ where $\widetilde{\pi} \in \Pi(\pi)$ and $\Pi(\pi)$ is the uncertainty set of the policy execution. Denote the actual behavior action at episode $k$ and step $h$ by $\widetilde{a}_{h}^{k}$ where $\widetilde{a}_{h}^{k} \sim \widetilde{\pi}_{h}^{k}$. Define the action robust value function of a policy $\pi$ as the worst-case expected accumulated reward over following any policy in the uncertainty set $\Pi(\pi)$ centered on a fixed policy $\pi$:
\begin{equation} \label{eq:robustV}
    V^{\pi}_{h}(s) = \min_{\widetilde{\pi} \in \Pi(\pi)} \mathbb{E}\left[ \sum_{h'=h}^{H} R_{h'}(s_{h'},a_{h'}) | s_h=s , a_{h'} \sim \widetilde{\pi}_{h'}(\cdot|s_{h'}), ~\forall h' > h \right].
\end{equation}
$V^{\pi}_{h}$ represents the action robust value function of policy $\pi$ at step $h$. Similarly, define the action robust $Q$-function of a policy $\pi$:
\begin{equation} \label{eq:robustQ}
    Q^{\pi}_{h}(s,a) = \min_{\widetilde{\pi} \in \Pi(\pi)} \mathbb{E}\left[ \sum_{h'=h}^{H} R_{h'}(s_{h'},a_{h'}) | s_h=s , a_h = a, a_{h'} \sim \widetilde{\pi}_{h'}(\cdot|s_{h'}), ~\forall h' > h \right].
\end{equation}
The goal of action robust RL is to find the optimal robust policy $\pi^*$ that maximizes the worst-case accumulated reward:
   $ \pi^* = \argmax_{\pi} V^{\pi}_{1}(s), \forall s \in \mathcal{S}.$
We also denote $V^{\pi^*}$ and $Q^{\pi^*}$ by $V^{*}$ and $Q^{*}$.

\paragraph{Probabilistic Policy Execution Uncertain Set} We follow the setting of the probabilistic action robust MDP (PR-MDP) introduced in \cite{tessler2019action} to construct the probabilistic policy execution uncertain set. For some  $0 \le \rho \le 1$, the policy execution uncertain set is defined as:
\begin{equation} \label{eq:piset}
      \Pi^{\rho}(\pi) :=\{\widetilde{\pi}| \forall s, \forall h, \widetilde{\pi}_h(\cdot|s) = (1-\rho) \pi_h(\cdot|s) + \rho \pi'_h(\cdot|s), \pi'_h(\cdot|s) \in \Delta_{\mathcal{A}}  \} = \otimes_{h,s}\Pi^{\rho}_{h,s}(\pi_h(\cdot|s))
\end{equation}
such that $\Pi^{\rho}_{h,s}(\pi_h(\cdot|s)) = \{\widetilde{\pi}_h(\cdot|s)| \widetilde{\pi}_h(\cdot|s) = (1-\rho) \pi_h(\cdot|s) + \rho \pi'_h(\cdot|s), \pi'_h(\cdot|s) \in \Delta_{\mathcal{A}}  \}$.

In this setting, an optimal probabilistic robust policy is optimal w.r.t. a scenario in which, with probability at most $\rho$, an adversary takes control and performs the worst possible action. We call $\pi'$ as the adversarial policy. \textcolor{black}{For different agent's policy $\pi$, the corresponding adversarial policy $\pi'$ that minimizes the cumulative reward may be different.}

The probabilistic uncertain set model is closely related to the uncertainty set models defined on the total variation distance. The uncertainty set based on distance is defined as $ \Pi^{D, \rho }(\pi) := \otimes_{h,s}\Pi^{D,\rho }_{h,s}(\pi_h(\cdot|s))$ such that $ \Pi^{D,\rho }_{h,s}(\pi_h(\cdot|s)) = \{\widetilde{\pi}_h(\cdot|s) \in \Delta_{\mathcal{A}} | D(\pi_h(\cdot|s), \widetilde{\pi}_h(\cdot|s)) \le \rho \}$, where $ D$ is some distance metric between two probability measures and $\rho$ is the radius. For any policy $\widetilde{\pi}_h(\cdot|s) \in \Pi^{\rho}_{h,s}(\pi_h(\cdot|s))$, the total variation distance to the center satisfies $D_{TV}(\pi_h(\cdot|s), \widetilde{\pi}_h(\cdot|s)) = \frac{1}{2}\lVert\pi_h(\cdot|s) - \widetilde{\pi}_h(\cdot|s)\rVert_1 \le \rho$.

\paragraph{Additional Notations} We set $\iota = \log(2SAHK/\delta)$ for $\delta > 0$. For simplicity of notation, we treat $P$ as a linear operator such that $[{P}_{h}V](s, a) := \mathbbm{E}_{s' \sim P_h(\cdot|s,a)} V(s')$, and we define two additional operators $\mathbbm{D}$ and $\mathbbm{V}$ as follows:
$[\mathbbm{D}_{\pi_h}Q](s) := \mathbbm{E}_{a \sim \pi_{h}(\cdot|s)} Q(s,a)$ and $\mathbbm{V}_{{P}_h}V_{h+1}(s,a) := \sum_{s'}{P}_h(s'|s,a)\left( V_{h+1}(s')- [{P}_hV_{h+1}](s,a) \right)^2 = [{P}_h(V_{h+1})^2](s,a) - ([{P}_hV_{h+1}](s,a))^2$. 

\section{Existence of the optimal robust policy}
For the standard tabular MDPs, when the state space, action space, and the horizon are all finite, there always exists an optimal policy. In addition, if the reward functions and the transition probabilities are known to the agent, the optimal policy can be solved by solving the Bellman optimality equation. In the following theorem, we show that the optimal policy also always exists in action robust MDPs and can be solved by the action robust Bellman optimality equation. 

\begin{theorem} \label{thm:Bellman}
If the uncertainty set of the policy execution has the form in \eqref{eq:piset}, the following perfect duality holds for all $s \in \mathcal{S}$ and all $h \in [H]$:
\begin{equation} \label{eq:duality}
\begin{split}
     &  \max_{\pi}  \min_{\widetilde{\pi} \in \Pi^{\rho}(\pi)} \mathbb{E}\left[ \sum_{h'=h}^{H} R_{h'}(s_{h'},a_{h'}) | s_h=s , a_{h'} \sim \widetilde{\pi}_{h'}(\cdot|s_{h'}) \right] \\
    =&    \min_{\widetilde{\pi} \in \Pi^{\rho}(\pi)} \max_{\pi}  \mathbb{E}\left[ \sum_{h'=h}^{H} R_{h'}(s_{h'},a_{h'}) | s_h=s , a_{h'} \sim \widetilde{\pi}_{h'}(\cdot|s_{h'}) \right]. 
\end{split}
\end{equation}
There always exists a deterministic optimal robust policy $\pi^*$. The problem can be solved via the induction of the action robust Bellman optimality equation on $h = H, \cdots, 1$. The action robust Bellman equation and the action robust Bellman optimality equation are:
\begin{equation} 
    \begin{split}
        \left\{
            \begin{aligned}
            & V^{\pi}_{h}(s)  = (1-\rho) [\mathbbm{D}_{\pi_h}Q^{\pi}_{h}](s) + \rho\min_{a\in\mathcal{A}}Q^{\pi}_{h}(s,a) ~\\
            & Q^{\pi}_{h}(s,a) =  R_h(s,a) + [{P}_{h}V_{h+1}^{\pi}](s,a)\\
            & V^{\pi}_{H+1}(s) = 0, ~ \forall s \in \mathcal{S}
            \end{aligned}
            \right.
    \end{split}
\end{equation}
\begin{equation} 
    \begin{split}
         \left\{
            \begin{aligned}
            & V^{*}_{h}(s) = (1-\rho)\max_{a\in\mathcal{A}}Q^{*}_{h}(s,a) + \rho\min_{b\in\mathcal{A}}Q^{*}_{h}(s,b)~\\
            & Q^{*}_{h}(s,a) =  R_h(s,a) + [{P}_{h}V_{h+1}^{*}](s,a)\\
            & V^{*}_{H+1}(s) = 0, ~ \forall s \in \mathcal{S}
            \end{aligned}
            \right..
    \end{split}
\end{equation}
\end{theorem}

Similar result of the perfect duality was show in \cite{tessler2019action}. They considered a PR-MDP as a two-player zero-sum Markov game and solving the optimal probabilistic robust policy can be equivalently viewed as solving the equilibrium value of a two-player zero-sum Markov game.
We define $C_h^{\pi,\pi',\rho}(s) := \mathbb{E}\left[ \sum_{h'=h}^{H} R_{h'}(s_{h'},a_{h'}) | s_h=s , a_{h'} \sim \widetilde{\pi}_{h'}(\cdot|s_{h'}) \right]$. The perfect duality of the control problems in ~\eqref{eq:duality} is equivalent to $ \max_{\pi}  \min_{\pi'} C_h^{\pi,\pi',\rho}(s) =   \min_{\pi'} \max_{\pi} C_h^{\pi,\pi',\rho}(s)$. We provide an alternate proof in Appendix~\ref{sec:ProofBellman} based on the robust Bellman equation.

\section{Algorithm and main results}
In this section, we introduce the proposed \textbf{A}ction \textbf{R}obust \textbf{R}einforcement \textbf{L}earning with \textbf{C}ertificates (ARRLC) algorithm and provides its theoretical guarantee. The pseudo code is listed in Algorithm~\ref{alg:ARRLC}. \textcolor{black}{Here, we highlight the main idea of our algorithm. Algorithm~\ref{alg:ARRLC} trains the agent in a clean (simulation) environment and learns a policy that performs well when applied to a perturbed environment with probabilistic policy execution uncertainty. To simulate the action perturbation, Algorithm~\ref{alg:ARRLC} chooses an adversarial action with probability $\rho$. To learn the agent's optimal policy and the corresponding adversarial policy, Algorithm~\ref{alg:ARRLC} computes an optimistic estimate $\overline{Q}$ of $Q^*$ and a pessimistic estimate $\underline{Q}$ of $Q^{\overline{\pi}^k}$. Algorithm~\ref{alg:ARRLC} uses the optimistic estimates to explore the possible optimal policy $\overline{\pi}$ and uses the pessimistic estimates to explore the possible adversarial policy $\underline{\pi}$. As shown later in Lemma~\ref{lem:Monot_mb}, $\overline{V} \ge V^{*} \ge V^{\overline{\pi}} \ge \underline{V}$ holds with high probabilities. The optimistic and pessimistic estimates $\overline{V}$ and $\underline{V}$ can provide policy certificates, which bounds the cumulative rewards of the return policy $\overline{\pi}^k$ and $\overline{V} - \underline{V}$ bounds the sub-optimality of the return policy $\overline{\pi}^k$ with high probabilities. The policy certificates can give us some insights about the performance of $\overline{\pi}^k$ in the perturbed environment with probabilistic policy execution uncertainty.}

\subsection{Algorithm  description}
We now describe the proposed ARRLC algorithm in more details. In each episode, the ARRLC algorithm can be decomposed into two parts.
\begin{itemize}

\item Line 3-11 (Sample trajectory and update the model estimate): Simulates the action robust MDP, executes the behavior policy $\widetilde{\pi}$, collects samples, and updates the estimate of the reward and the transition.

\item  Line 16-25 (Adversarial planning from the estimated model):  Performs value iteration with bonus to estimate the robust value functions using the empirical estimate of the transition $\hat{P}$, computes a new policy $\overline{\pi}$ which is optimal respect to the estimated robust value functions, and computes a new optimal adversarial policy $\underline{\pi}$ respect to the agent's policy $\overline{\pi}$.
\end{itemize}

At a high-level, this two-phase strategy is standard in the majority of model-based RL algorithms~\cite{UCBVI,dann2019policy}. Algorithm~\ref{alg:ARRLC} shares similar structure with ORLC (Optimistic Reinforcement Learning with Certificates) in \cite{dann2019policy} but has some significant differences in line 5-6 and line 18-23. The first main difference is that the ARRLC algorithm simulates the probabilistic policy execution uncertainty by choosing an adversarial action with probability $\rho$. The adversarial policy and the adversarial action are computed by the ARRLC algorithm. The second main difference is that the ARRLC algorithm simultaneously plans the agent policy $\overline{\pi}$ and the adversarial policy $\underline{\pi}$ by the action robust Bellman optimality equation.

These two main difference brings two main challenges in the design and analysis of our algorithm.

(1) The ARRLC algorithm simultaneously plans the agent policy and the adversarial policy. However the planned adversarial policy $\underline{\pi}$ is not necessarily the true optimal adversary policy towards the agent policy $\overline{\pi}$ because of the estimation error of the value functions. We carefully design the bonus items and the update role of the value functions so that $\overline{V}_h (s) \ge V_h^*(s) \ge V_h^{\overline{\pi}}(s) \ge \underline{V}_h (s)$ and $\overline{Q}_h (s,a) \ge Q_h^*(s,a) \ge Q_h^{\overline{\pi}}(s,a) \ge \underline{Q}_h (s,a)$ hold for all $s$ and $a$.

(2) A crucial step in many UCB-type algorithms based on Bernstein inequality is bounding the sum of variance of estimated value function across the planning horizon. The behavior policies in these UCB-type algorithms are deterministic. However, the behavior policy in our ARRLC algorithm is not deterministic due to the simulation of the adversary's behavior. The total variance is the weighted sum of the sum of variance of estimated value function across two trajectories. Even if action $\overline{\pi}(s_h^k)$ or $\underline{\pi}(s_h^k)$ is not sampled at state $s_h^k$, it counts in the total variance.

\begin{algorithm}[htb] 
	\caption{ARRLC (\textbf{A}ction \textbf{R}obust \textbf{R}einforcement \textbf{L}earning with \textbf{C}ertificates)}   	\label{alg:ARRLC} 
	\begin{algorithmic}[1] 
		\STATE  Initialize $\overline{V}_{h}(s) = H-h+1$, $\overline{Q}_{h}(s,a) = H-h+1$, $\underline{V}_{h}(s) = 0$, $\underline{Q}_{h}(s,a) = 0$, $\hat{r}_h(s,a)$, $N_h(s,a) = 0$ and $N_h(s,a, s') = 0$ for all state $s \in \mathcal{S}$, all action $a \in \mathcal{A}$ and all step $h \in [H]$.  $\overline{V}_{H+1}(s)=\underline{V}_{H+1}(s) =0 $  and $\overline{Q}_{H+1}(s,a)=\underline{Q}_{H+1}(s,a) = 0$ for all $s$ and $a$. $\Delta = H$.
		\FOR{episode $k = 1, 2, \dots,K$}
		\FOR{step $h = 1, 2, \dots, H$} 
		\STATE Observe $s_h^k$. 
		\STATE Set $\overline{\pi}_h^k(s) = \argmax_a \overline{Q}_h(s,a) $ , $\underline{\pi}_h^k(s) = \argmin_a \underline{Q}_{h}(s,a)$, $\widetilde{\pi}_h^k = (1-\rho)\overline{\pi}_h^k +\rho\underline{\pi}_h^k$.
		\STATE Take action $a_h^k \sim \widetilde{\pi}_h^k(\cdot|s_h^k) $.
		\STATE Receive reward $r_h^k$ and observe $s_{h+1}^k$.
		\STATE Set $N_h(s_h^k,a_h^k) \leftarrow N_h(s_h^k,a_h^k)+1$, $N_h(s_h^k,a_h^k,s_{h+1}^k) \leftarrow N_h(s_h^k,a_h^k,s_{h+1}^k)+1$.
		\STATE Set $\hat{r}_h^k(s_h^k,a_h^k) \leftarrow \hat{r}_h^k(s_h^k,a_h^k)+(r_h^k - \hat{r}_h^k(s_h^k,a_h^k))/ N_h(s_h^k,a_h^k)$.
		\STATE Set $\hat{P}_h(\cdot|s_h^k,a_h^k) = N_h(s_h^k,a_h^k,\cdot) / N_h(s_h^k,a_h^k)$.
		\ENDFOR
		\STATE \textbf{Output} policy $\overline{\pi}^k$ with certificates $\mathcal{I}_k = [ \underline{V}_1(s_1^k), \overline{V}_1(s_1^k) ]$ and $\epsilon_k= |\mathcal{I}_k|$ .
		\IF{$\epsilon_k < \Delta$}
		\STATE $\Delta \leftarrow \epsilon_k$ and $\pi^{out} \leftarrow \overline{\pi}^k$. 
		\ENDIF
		\FOR{step $h = H, H-1, \dots, 1$ }  
		\FOR{each $(s, a) \in \mathcal{S} \times \mathcal{A}$ with $N_h(s,a) > 0$} 
		\STATE Set $\theta_h(s,a) = \sqrt{\frac{2\mathbbm{V}_{\hat{P}_h}[(\overline{V}_{h+1}+\underline{V}_{h+1})/2](s,a)\iota }{N_h(s,a)  }} +  \sqrt{\frac{2 \hat{r}_h(s,a) \iota}{N_h(s,a)  }} + \frac{  \hat{P}_h \left(\overline{V}_{h+1}-\underline{V}_{h+1} \right)(s,a)}{H} + \frac{ (24H^2+7H+7) \iota}{3N_h(s,a) }$,
		\STATE $\overline{Q}_h(s,a) \leftarrow \min\{H-h+1, \hat{r}_h(s,a) + \hat{P}_h\overline{V}_{h+1}(s,a) + \theta_h(s,a) \}$,
		\STATE $\underline{Q}_h(s,a) \leftarrow \max\{0, \hat{r}_h(s,a) + \hat{P}_h\underline{V}_{h+1}(s,a) - \theta_h(s,a) \}$,
		\STATE $\overline{\pi}_h^{k+1}(s) = \argmax_a \overline{Q}_h(s,a) $ , $\underline{\pi}_h^{k+1}(s) = \argmin_a \underline{Q}_{h}(s,a)$,
		\STATE $\overline{V}_h(s) \leftarrow (1-\rho)\overline{Q}_h(s, \overline{\pi}_h^{k+1}(s)) + \rho \overline{Q}_h(s, \underline{\pi}_h^{k+1}(s)) $,
		\STATE $\underline{V}_h(s) \leftarrow (1-\rho)\underline{Q}_h(s, \overline{\pi}_h^{k+1}(s)) + \rho \underline{Q}_h(s, \underline{\pi}_h^{k+1}(s)) $.
		\ENDFOR
		\ENDFOR
		\ENDFOR
		\RETURN $\pi^{out}$
	\end{algorithmic}
\end{algorithm}

\subsection{Theoretical guarantee}
We define the cumulative regret of the output policy $\overline{\pi}^k$ at each episodes $k$ as $Regret(K) := \sum_{k=1}^K ( V_{1}^*(s_1^k) -  {V}_1^{\overline{\pi}^k}(s_1^k))$.
\begin{theorem} \label{thm:ARRLC}
For any $\delta \in (0, 1]$, letting  $\iota = \log(2SAHK/\delta)$, then with
probability at least $1 - \delta$, Algorithm~\ref{alg:ARRLC} achieves:
\begin{itemize}
    \item $V_1^*(s_1) -V_1^{\pi^{out}}(s_1) \le \epsilon$, if the number of episodes $K \ge \Omega(SAH^3\iota^2/\epsilon^2 + S^2AH^3\iota^2/\epsilon)$.
    \item $Regret(K) = \sum_{k=1}^K ( V_{1}^*(s_1^k) -  {V}_1^{\overline{\pi}^k}(s_1^k)) \le \mathcal{O}(\sqrt{SAH^3K}\iota + S^2AH^3\iota^2)$.
\end{itemize}
\end{theorem}

For small $\epsilon \le H/S$, the sample complexity scales as $\mathcal{O}(SAH^3\iota^2/\epsilon^2)$. For the case with a large number of episodes $K \ge S^3AH^3\iota$, the regret scales as $\mathcal{O}(\sqrt{SAH^3K}\iota) $.
For the standard MDPs, the information-theoretic sample complexity lower bound is $\Omega(SAH^3/\epsilon^2)$ provided in ~\cite{zhang2020almost} and the regret lower bound is $\Omega(\sqrt{SAH^3K})$ provided in ~\cite{jin2018q}. When $\rho = 0$, the action robust MDPs is equivalent to the standard MDPs. Thus, the information-theoretic sample complexity lower bound and the regret lower bound of the action robust MDPs should have same dependency on $S$, $A$, $H$, $K$ or $\epsilon$. The lower bounds show the optimality of our algorithm up to logarithmic factors.

\section{Proof sketch }

In this section, we provide sketch of the proof, which will highlight our the main ideas of our proof. First, we will show that $\overline{V}_h (s) \ge V_h^*(s) \ge V_h^{\overline{\pi}}(s) \ge \underline{V}_h (s)$ hold for all $s$ and $a$. Then, the regret can be bounded by $\overline{V}_1 - \underline{V}_1$ and then be divided by four items, each of which can then be bounded separately. The full proof can be found in the appendix contained in the supplementary material.

We first introduce a few notations. We use 
$\overline{Q}_h^k$,$\overline{V}_h^k$,$\underline{Q}_h^k$,$\underline{V}_h^k$, $N_h^k$, $\hat{P}_h^k$,$\hat{r}_h^k$ and $\theta_h^k$ to denote the values of $\overline{Q}_h$,$\overline{V}_h$,$\underline{Q}_h$,$\underline{V}_h$, $\max\{N_h,1\}$, $\hat{P}_h$, ${r}_h$ and $\theta_h$ in the beginning of the $k$-th episode in Algorithm~\ref{alg:ARRLC}.

\subsection{Proof of monotonicity}
We define $\mathcal{E}^R$ to be the event where 
\begin{equation} \label{eq:eventR}
    \left|\hat{r}_h^k(s,a) - R_h(s,a) \right| \le \sqrt{\frac{2 \hat{r}_h^k(s,a) \iota}{N_h^k(s,a)  }} + \frac{7 \iota}{3(N_h^k(s,a) )}
\end{equation}
holds for all $(s, a, h, k) \in S \times A \times [H] \times[K]$.
We also define $\mathcal{E}^{PV}$ to be the event where 
\begin{eqnarray} 
 \left| (\hat{P}_h^k - P_h)V^*_{h+1}(s,a)  \right| &\le& \sqrt{\frac{2\mathbbm{V}_{\hat{P}_h^k}V_{h+1}^*(s,a) \iota}{N_h^k(s,a)  }} + \frac{7H \iota}{3(N_h^k(s,a) )}\label{eq:eventPV1}\\
 \left| (\hat{P}_h^k - P_h)V^{\overline{\pi}^k}_{h+1}(s,a)  \right| &\le& \sqrt{\frac{2\mathbbm{V}_{\hat{P}_h^k}V_{h+1}^{\overline{\pi}^k}(s,a) \iota}{N_h^k(s,a)  }} + \frac{7H \iota}{3N_h^k(s,a) } \label{eq:eventPV2}
\end{eqnarray}
holds for all $(s, a, h, k) \in S \times A \times [H] \times[K]$. 

Event $\mathcal{E}^R$ means that the estimations of all reward functions stay in certain neighborhood of the true values. Event $\mathcal{E}^{PV}$ represents that the estimation of the value functions at the next step stay in some intervals. The following lemma shows $\mathcal{E}^R$ and $\mathcal{E}^{PV}$ hold with high probability. The analysis will be done assuming the successful event $\mathcal{E}^R \cap \mathcal{E}^{PV}$ holds in the rest of this section.

\begin{lemma} \label{lem:event_mb}
$\mathbbm{P}(\mathcal{E}^R \cap \mathcal{E}^{PV}) \ge 1- 3\delta$.
\end{lemma}

\begin{lemma} \label{lem:Monot_mb}
Conditioned on $\mathcal{E}^R \cap \mathcal{E}^{PV}$, $\overline{V}_h^k (s) \ge V_h^*(s) \ge V_h^{\overline{\pi}^k}(s) \ge \underline{V}_h^k (s)$ and $\overline{Q}_h^k (s,a) \ge Q_h^*(s,a) \ge Q_h^{\overline{\pi}^k}(s,a) \ge \underline{Q}_h^k (s,a)$ for all $(s, a, h, k) \in S \times A \times [H] \times[K]$.
\end{lemma}

\subsection{Regret analysis}
We decompose the regret and analyze the different terms.  Set  $\Theta_h^k(s,a) = \sqrt{\frac{8\mathbbm{V}_{{P}_h}{C}^{{\pi}^{k*},\underline{\pi}^k,\rho}_{h+1}(s,a)\iota }{N_h^k(s,a)  }} + \sqrt{\frac{32}{N_h^k(s,a)  } } + \frac{46 \sqrt{SH^4\iota}}{N_h^k(s,a)  } $, where ${\pi}^{k*}$ is the optimal policy towards the adversary policy $\underline{\pi}^k$ with ${\pi}^{k*}_{h}(s) = \argmax_\pi {C}^{{\pi},\underline{\pi}^k,\rho}_{h}(s)$. We define the cumulative regret of the output policy $\overline{\pi}^k$ at each episodes $k$ as $Regret(K) := \sum_{k=1}^K ( V_{1}^*(s_1^k) -  {V}_1^{\overline{\pi}^k}(s_1^k))$. Let $M_1 = \sum_{k=1}^K \sum_{h=1}^H [\mathbbm{D}_{\widetilde{\pi}_h^k}\hat{P}_h^k(\overline{V}^k_{h+1} - \underline{V}^k_{h+1})(s_h^k)- \hat{P}_h^k(\overline{V}_{h+1}^k - \underline{V}_{h+1}^k)(s_h^k,a_h^k)]$, $M_2 =  \sum_{k=1}^K \sum_{h=1}^H \frac{1}{H}[\mathbbm{D}_{\widetilde{\pi}_h^k}P_h(\overline{V}^k_{h+1} - \underline{V}^k_{h+1})(s_h^k)- {P}_h(\overline{V}_{h+1}^k - \underline{V}_{h+1}^k)(s_h^k,a_h^k) ] $, $M_3 = \sum_{k=1}^K \sum_{h=1}^H ({P}_h^k(\overline{V}_{h+1}^k - \underline{V}_{h+1}^k)(s_h^k,a_h^k) - (\overline{V}_{h+1}^k - \underline{V}_{h+1}^k)(s_{h+1}^k) )  $ and $M_4 = \sum_{k=1}^K \sum_{h=1}^H [\frac{(SH+SH^2)\iota}{N_h^k(s_h^k,a_h^k)}+\mathbbm{D}_{\widetilde{\pi}_h^k}\Theta_h^k(s_h^k)] $. \textcolor{black}{Here $M_1$ and $M_2$ are the cumulative sample error from the random choices of the adversarial policy or agent's policy. $M_3$ is the cumulative sample error from the randomness of Monte Carlo sampling of the next state. $M_4$ is the cumulative error from the bonus item $\theta$. Lemma~\ref{lem:regdec} shows that the regret can be bounded by these four terms.}

\begin{lemma} \label{lem:regdec}
With probability at least $1 -(S+5)\delta$, 
\begin{equation}
	Regret(K) \le  \sum_{k=1}^K (\overline{V}_{1}^k(s_1^k) -  \underline{V}_1^{k}(s_1^k)) 
	\le   21 (M_1+M_2+M_3+M_4).
\end{equation}


\end{lemma}

We now bound each of these four items separately. 
\begin{lemma} \label{lem:M1}
With probability at least $1 - \delta$, $ |M_1| \le H\sqrt{2HK\iota}. $
\end{lemma}

\begin{lemma}  \label{lem:M2}
With probability at least $1 -\delta$, $|M_2| \le \sqrt{2HK\iota}$.
\end{lemma}

\begin{lemma} \label{lem:M3}
With probability at least $1 - \delta$, $ |M_3| \le  H\sqrt{2HK\iota}.$
\end{lemma}

\begin{lemma}  \label{lem:M4}
With probability at least $1 -2\delta$, $|M_4| \le 2S^2AH^3\iota^2 + 8\sqrt{SAH^2K\iota} +46S^{\frac{3}{2}}AH^3\iota^2 +\sqrt{24SAH^3K}\iota +  6\sqrt{SAH^5}\iota $.
\end{lemma}

\paragraph{Putting All Together} By Lemmas~\ref{lem:regdec},~\ref{lem:M1},~\ref{lem:M2},~\ref{lem:M3}, and~\ref{lem:M4}, we conclude that, with probability $1-(S+10)\delta$,
\begin{equation}
\begin{split}
    Regret(K) \le & O(\sqrt{H^3K\iota} + \sqrt{SAH^2K\iota} + \sqrt{SAH^3K}\iota + S^2AH^3\iota^2 + \sqrt{SAH^5}\iota ) \\
     =& O(\sqrt{SAH^3K}\iota + S^2AH^3\iota^2).
\end{split}
\end{equation}
By rescaling $\delta$, $\log(\frac{2SAHK}{\delta/(S+10)}) \le c\iota $ for some constant $c$ and we finish the proof of regret. As $\sum_{k=1}^K (\overline{V}_{1}^k(s_1^k) -  \underline{V}_1^{k}(s_1^k)) \le O(\sqrt{SAH^3K}\iota + S^2AH^3\iota^2)$, we have that $V_{1}^*(s_1) -  {V}_1^{{\pi}^{out}}(s_1)\le \min_k \overline{V}_{1}^k(s_1^k) -  \underline{V}_1^{k}(s_1^k) \le O(\frac{\sqrt{SAH^3}\iota}{K} + \frac{S^2AH^3\iota^2}{K})$ and we finish the proof of sample complexity.

\section{Model-free method}
In this section, we develop a model-free algorithm and analyze its theoretical guarantee.
We present the proposed Action Robust Q-learning with UCB-Hoeffding (AR-UCBH) algorithm show in Algorithm~\ref{alg:AR-UCBH}. Here, we highlight the main idea of Algorithm~\ref{alg:AR-UCBH}. Algorithm~\ref{alg:AR-UCBH} follows the same idea of  Algorithm~\ref{alg:ARRLC}, which trains the agent in a clean (simulation) environment and learns a policy that performs well when applied to a perturbed environment with probabilistic policy execution uncertainty. To simulate the action perturbation, Algorithm~\ref{alg:AR-UCBH} chooses an adversarial action with probability $\rho$. To learn the agent's optimal policy and the corresponding adversarial policy, Algorithm~\ref{alg:AR-UCBH} computes an optimistic estimate $\overline{Q}$ of $Q^*$ and a pessimistic estimate $\underline{Q}$ of $Q^{\overline{\pi}^k}$. Algorithm~\ref{alg:AR-UCBH} uses the optimistic estimates to explore the possible optimal policy $\overline{\pi}$ and uses the pessimistic estimates to explore the possible adversarial policy $\underline{\pi}$. The difference is that Algorithm~\ref{alg:AR-UCBH} use a model-free method to update $Q$ and $V$ values.

\begin{algorithm}[htb] 
 	\caption{Action Robust Q-learning with UCB-Hoeffding (AR-UCBH)}   	\label{alg:AR-UCBH} 
 	\begin{algorithmic}[1] 
 	     \STATE Set $\alpha_t = \frac{H+1}{H+t}$. Initialize $\overline{V}_{h}(s) = H-h+1$, $\overline{Q}_{h}(s,a) = H-h+1$, $\underline{V}_{h}(s) = 0$, $\underline{Q}_{h}(s,a) = 0$, $\hat{r}_h(s,a)$, $N_h(s,a) = 0$  for all state $s \in \mathcal{S}$, all action $a \in \mathcal{A}$ and all step $h \in [H]$.  $\overline{V}_{H+1}(s)=\underline{V}_{H+1}(s) =0 $  and $\overline{Q}_{H+1}(s,a)=\underline{Q}_{H+1}(s,a) = 0$ for all $s$ and $a$. $\Delta = H$. Initial policy $\overline{\pi}_h^1(a|s)$ and $\underline{\pi}_h^1(a|s) = 1/A$ for all state $s$, action $a$ and all step $h \in [H]$.
 		\FOR{episode $k = 1, 2, \dots,K$}
 		\FOR{step $h = 1, 2, \dots, H$}
    		\STATE Observe $s_h^k$. 
            \STATE Set  $\overline{a}_h^k = \argmax_a \overline{Q}_h(s_h^k,a) $ , $\underline{a}_h^k = \argmin_a \underline{Q}_{h}(s_h^k,a)$,  $\widetilde{\pi}_h^k(\overline{a}_h^k |s_h^k) = 1-\rho$ and $\widetilde{\pi}_h^k(\underline{a}_h^k |s_h^k) = \rho$.
 		\STATE Take action $a_h^k \sim \widetilde{\pi}_h^k(\cdot|s_h^k) $.
 		\STATE Receive reward $r_h^k$ and observe $s_{h+1}^k$.
 		\STATE Set $t = N_h(s_h^k,a_h^k) \leftarrow N_h(s_h^k,a_h^k)+1$; $b_t = \sqrt{H^3\iota/t}$.
\STATE $\overline{Q}_h(s_h^k,a_h^k) \leftarrow  (1-\alpha_t)\overline{Q}_h(s_h^k,a_h^k) + \alpha_t(r_h^k+\overline{V}_{h+1}(s_{h+1}^k)+b_t) $,
   \STATE $\underline{Q}_h(s_h^k,a_h^k) \leftarrow  (1-\alpha_t)\underline{Q}_h(s_h^k,a_h^k) + \alpha_t(r_h^k+\underline{V}_{h+1}(s_{h+1}^k)- b_t) $.
      \STATE Set $\overline{\pi}_h^{k+1}(s_h^k) = \argmax_a \overline{Q}_h(s_h^k,a)$, $\underline{\pi}_h^{k+1}(s_h^k) = \argmin_a \underline{Q}_h(s_h^{k+1},a) $. 
\STATE $\overline{V}_h(s_h^k) \leftarrow \min\{\overline{V}_h(s_h^k), (1-\rho)\overline{Q}_h(s_h^k, \overline{\pi}_h^{k+1}(s_h^k)) + \rho \overline{Q}_h(s_h^k, \underline{\pi}_h^{k+1}(s_h^k))\}$.
    \STATE $\underline{V}_h(s_h^k) \leftarrow \max\{\underline{V}_h(s_h^k) , (1-\rho)\underline{Q}_h(s_h^k, \overline{\pi}_h^{k+1}(s_h^k)) + \rho \underline{Q}_h(s_h^k, \underline{\pi}_h^{k+1}(s_h^k))\}$.
  \IF{$\underline{V}_h(s_h^k) > (1-\rho)\underline{Q}_h(s_h^k, \overline{\pi}_h^{k+1}(s_h^k)) + \rho \underline{Q}_h(s_h^k, \underline{\pi}_h^{k+1}(s_h^k)) $}   
   \STATE $\overline{\pi}_h^{k+1} = \overline{\pi}_h^{k}$.
   \ENDIF
        \ENDFOR
 \STATE \textbf{Output} policy $\overline{\pi}^{k+1}$ with certificates $\mathcal{I}_{k+1} = [ \underline{V}_1(s_1^k), \overline{V}_1(s_1^k) ]$ and $\epsilon_{k+1}= |\mathcal{I}_{k+1}|$.
   \ENDFOR
   \RETURN $\overline{\pi}^{k+1}$
 	\end{algorithmic}
 \end{algorithm}

Here, we highlight the challenges of the model-free planning compared with the model-based planing. In the model-based planning, we performs value iteration and the $Q$ values, $V$ values, agent policy $\overline{\pi}$ and adversarial policy $\underline{\pi}$ are updated on all $(s,a)$. However, in the model-free method, the $Q$ values, $V$ values are updated only on $(s_h^k, a_h^k)$ which are the samples on the trajectories. Compared with the model-based planning, the model-free planning is slower and less stable. We need to update the output policy carefully. In line 14-16, Algorithm~\ref{alg:AR-UCBH} does not update the output policy when the lower bound on the value function of the new policy does not improve. By this, the output policies are stably updated.

We provide the regret and sample complexity bounds of Algorithm~\ref{alg:AR-UCBH} in the following:
\begin{theorem} \label{thm:AR-UCBH}
For any $\delta \in (0, 1]$, letting  $\iota = \log(2SABHK/\delta)$, then with
probability at least $1 - \delta$, Algorithm~\ref{alg:AR-UCBH} achieves:
\begin{itemize}
    \item $V_1^*(s_1) -V_1^{\pi^{out}}(s_1) \le \epsilon$, if the number of episodes $K \ge \Omega(SAH^5\iota/\epsilon^2 + SAH^2/\epsilon)$.
    \item $Regret(K) = \sum_{k=1}^K ( V_{1}^*(s_1^k) -  {V}_1^{\overline{\pi}^k}(s_1^k)) \le \mathcal{O}(\sqrt{SAH^5K\iota} + SAH^2)$.
\end{itemize}
\end{theorem}

The detailed proof is provided in Appendix~\ref{sec:prfmf}
\section{Simulation results}

\begin{figure}[ht]
     \centering
    \subfigure[p=0.1,  fix]{
         \includegraphics[width=0.23\textwidth]{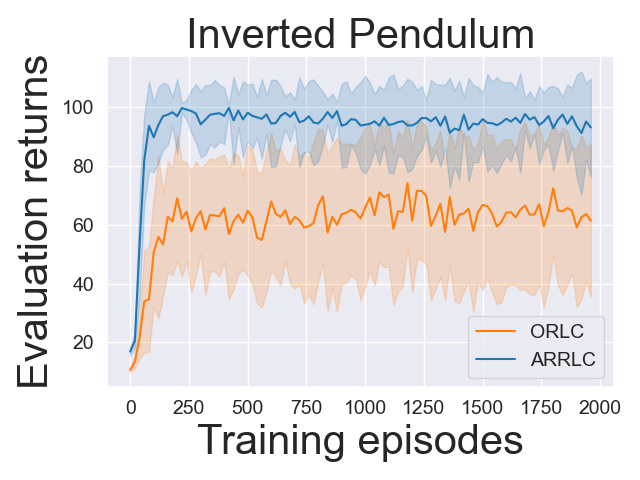}}
    \subfigure[p=0.2,  fix]{
         \includegraphics[width=0.23\textwidth]{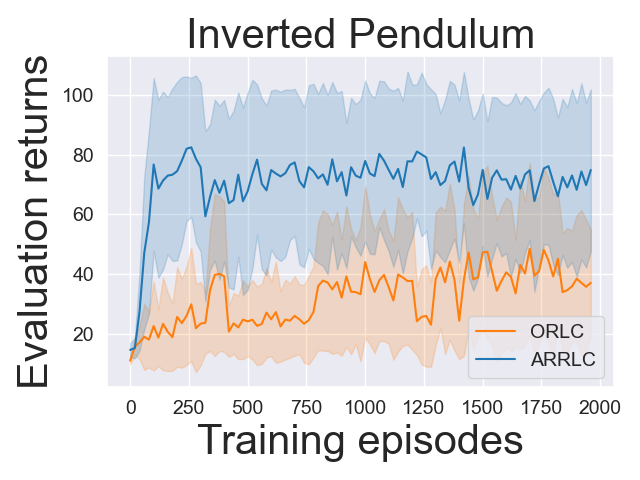}}
    \subfigure[p=0.1,  random]{
         \includegraphics[width=0.23\textwidth]{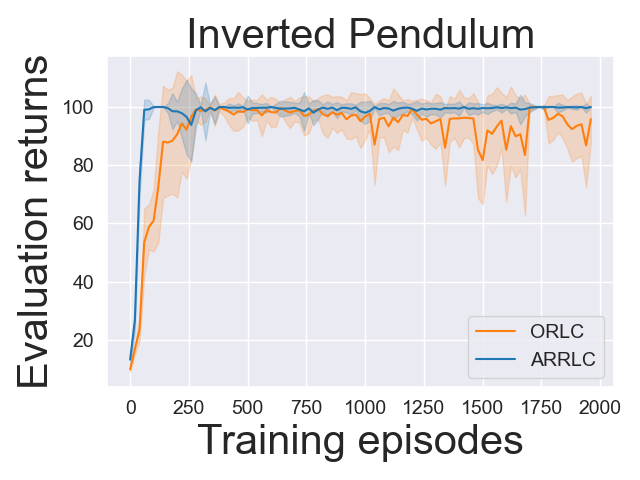}}
    \subfigure[p=0.2,  random]{
         \includegraphics[width=0.23\textwidth]{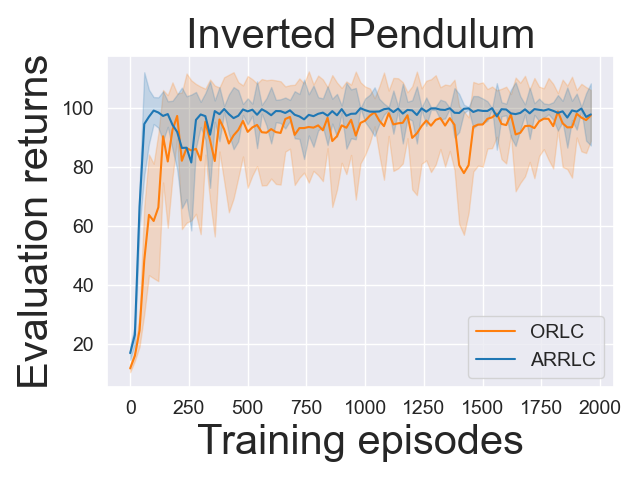}}   

    \subfigure[p=0.1,  fix]{
         \includegraphics[width=0.23\textwidth]{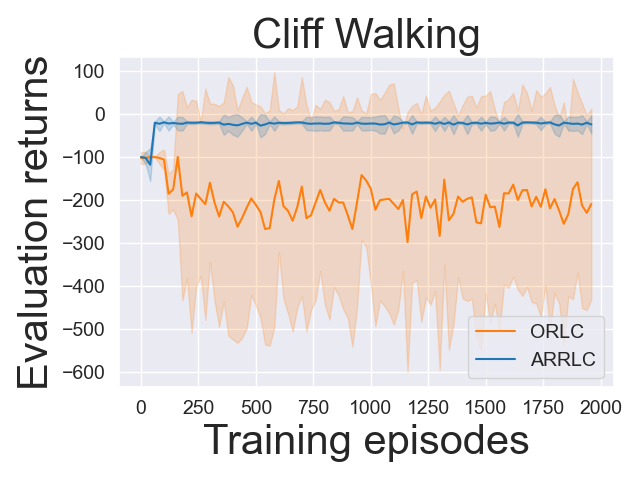}}
    \subfigure[p=0.2,  fix]{
         \includegraphics[width=0.23\textwidth]{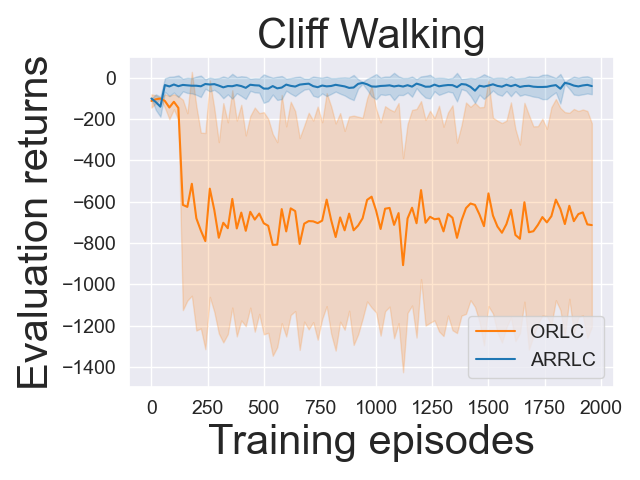}}
    \subfigure[p=0.1,  random]{
         \includegraphics[width=0.23\textwidth]{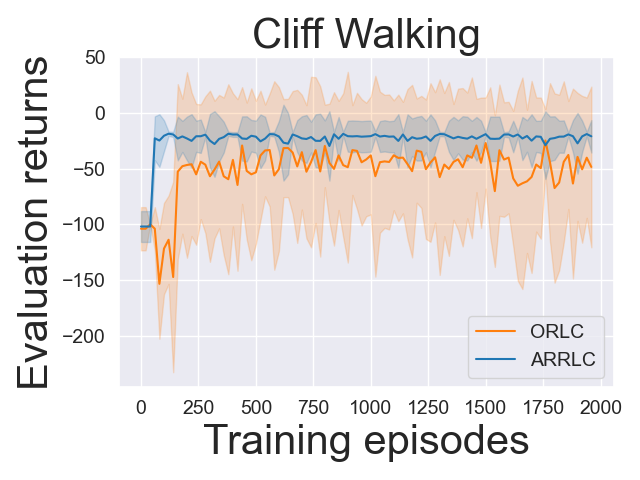}}
    \subfigure[p=0.2,  random]{
         \includegraphics[width=0.23\textwidth]{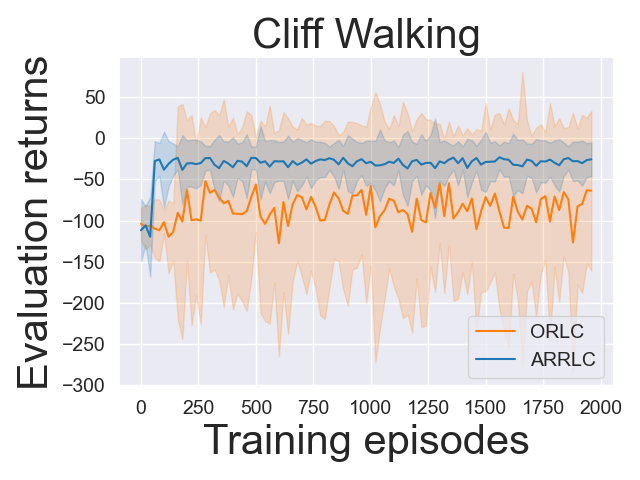}} 
        \caption{ARRLC v.s. ORLC.}
        \label{fig:ARRLCvsORLC}
\end{figure}

\begin{figure}[ht]
     \centering
    \subfigure[p=0.1,  fix]{
         \includegraphics[width=0.23\textwidth]{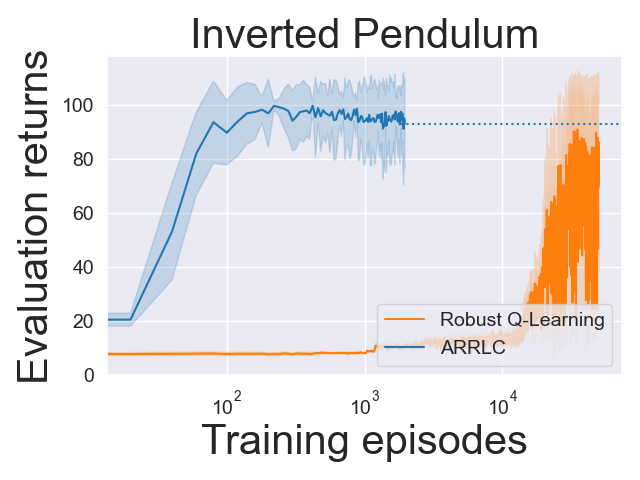}}
    \subfigure[p=0.2,  fix]{
         \includegraphics[width=0.23\textwidth]{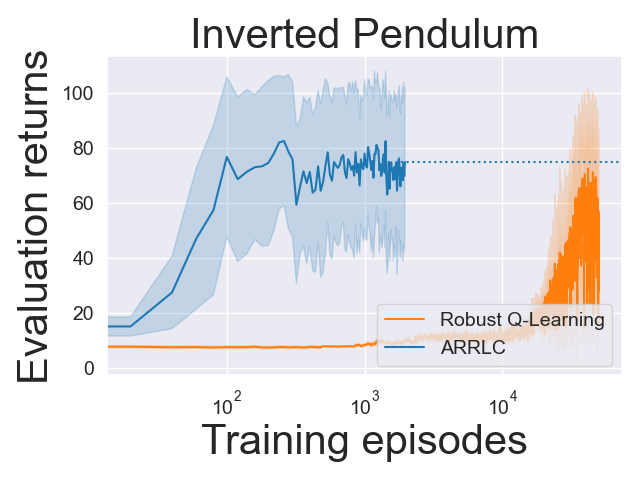}}
    \subfigure[p=0.1,  random]{
         \includegraphics[width=0.23\textwidth]{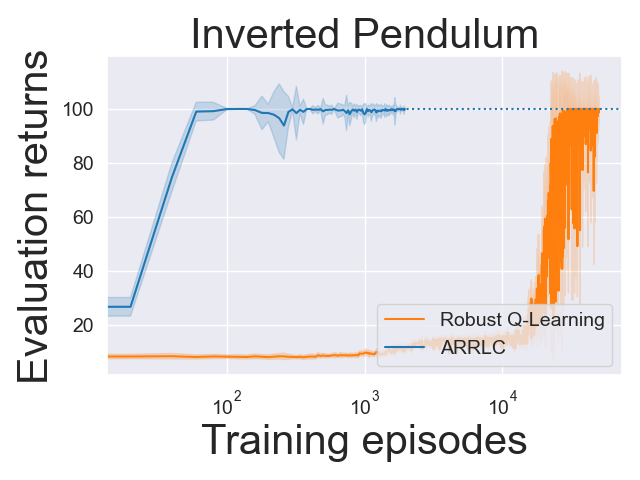}}
    \subfigure[p=0.2,  random]{
         \includegraphics[width=0.23\textwidth]{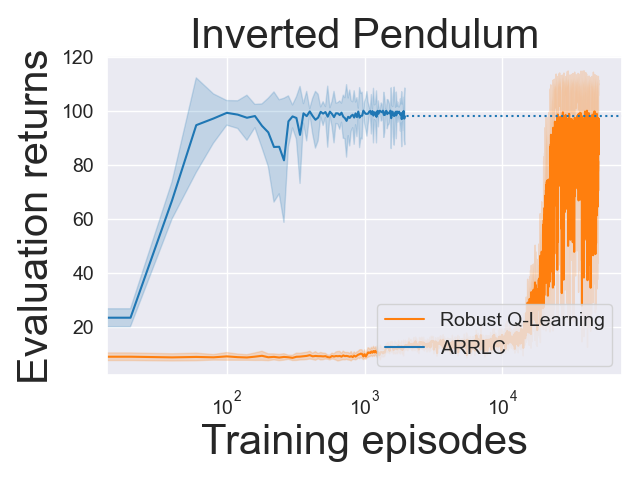}} 

    \subfigure[p=0.1,  fix]{
         \includegraphics[width=0.23\textwidth]{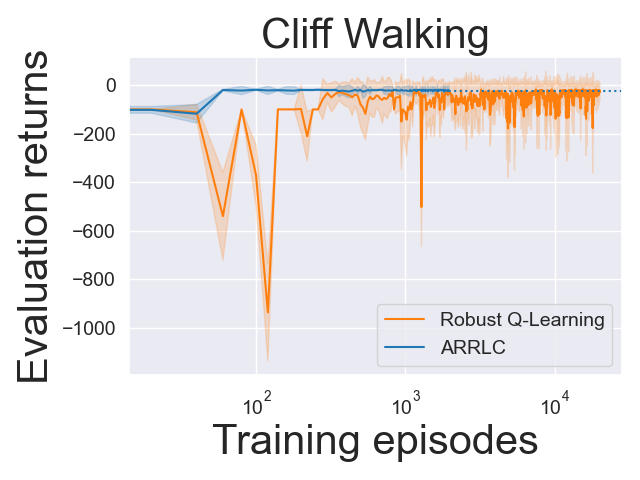}}
    \subfigure[p=0.2,  fix]{
         \includegraphics[width=0.23\textwidth]{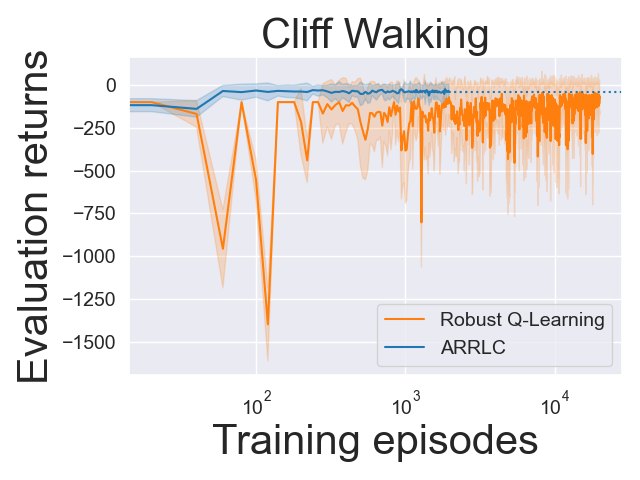}}
    \subfigure[p=0.1,  random]{
         \includegraphics[width=0.23\textwidth]{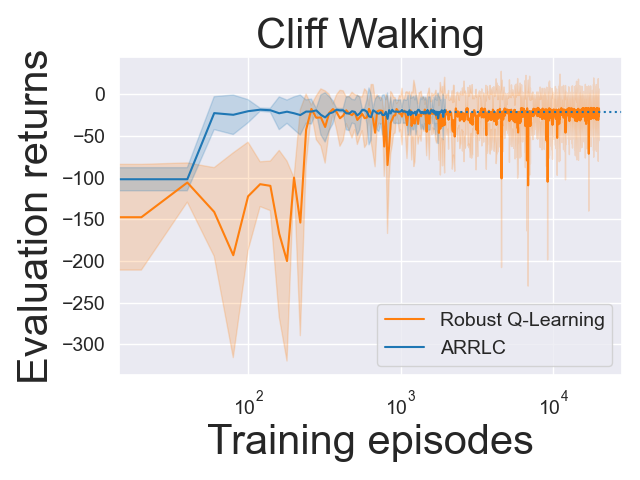}}
    \subfigure[p=0.2,  random]{
         \includegraphics[width=0.23\textwidth]{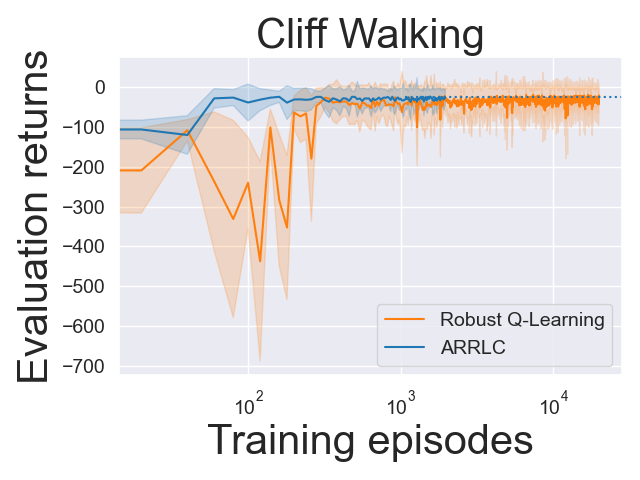}} 
        \caption{ARRLC v.s. Robust TD}
        \label{fig:ARRLCvsRobustTD}
\end{figure}

We use OpenAI gym framework \cite{brockman2016openai}, and consider two different problems: Cliff Walking, a toy text environment, and Inverted Pendulum, a control environment with the MuJoCo \cite{todorov2012mujoco} physics simulator. We set $H = 100$.  To demonstrate the robustness, the policy is learned in a clean environment, and is then tested on the perturbed environment.
Specifically, during the testing, we set a probability $p$ such that after the agent takes an action, with probability $p$, the action is uniformly randomly choosen or choosen by a fixed adversarial policy. A Monte-Carlo method is used to evaluate the accumulated reward of the learned policy on the perturbed environment. We take the average over 100 trajectories. Training of ARRLC on CliffWalking-v0 and InvertedPendulum-v4 respectively cost roughly 5 seconds and 22 seconds per 100 episodes on an i9-9880H CPU core.
\textcolor{black}{In Figure~\ref{fig:ARRLCvsORLC} and Figure~\ref{fig:ARRLCvsRobustTD}, "fix" represents that the action is perturbed by a fixed adversarial policy during the testing, "random" represents that the action is randomly perturbed during the testing, $p$ is the action perturbation probability. }

\begin{figure*}[ht] 
\centering
\begin{minipage}[t]{0.45\textwidth}
\centering
\includegraphics[width=0.85\textwidth]{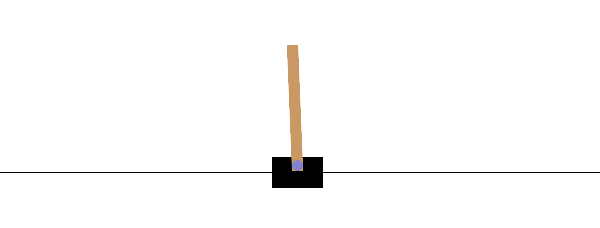}
\caption{Inverted pendulum environment.}
\label{fig:ip}
\end{minipage}
\begin{minipage}[t]{0.45\textwidth}
\centering
\includegraphics[width=0.85\textwidth]{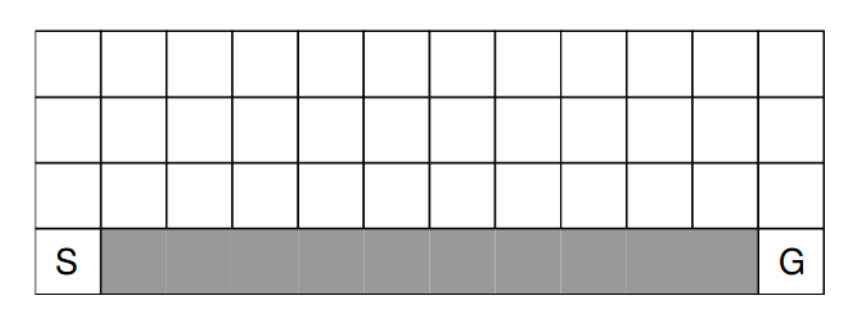} 
\caption{Cliff walking environment.}
\label{fig:cw}
\end{minipage}
\end{figure*}

\paragraph{Inverted pendulum}
The inverted pendulum experiment as shown in Figure~\ref{fig:ip} is a classic control problem in RL. An inverted pendulum is attached by a pivot point to a cart, which is restricted to linear movement in a plane. The cart can be pushed left or right, and the goal is to balance the  inverted pendulum on the top of the cart by applying forces on the cart. A reward of $+1$ is awarded for each time step that the inverted pendulum stand upright within a certain angle limit. The fixed adversarial policy in the inverted pendulum environment is a force of $0.5$ N in the left direction.

\paragraph{Cliff walking}
The cliff walking experiment as shown in Figure~\ref{fig:cw} is a classic scenario proposed in \cite{sutton2018reinforcement}. The game starts with the player at location $[3, 0]$ of the $4 \times 12$ grid world with the goal located at $[3, 11]$. A cliff runs along $[3, 1-10]$. If the player moves to a cliff location, it returns to the start location and receives a reward of $-100$. For every move which does not lead into the cliff, the agent receives a reward of $-1$. The player makes moves until they reach the goal. The fixed adversarial policy in the cliff walking environment is walking a step to the bottom.

We compare our algorithm with the non-robust RL algorithm, which is ORLC (Optimistic Reinforcement Learning with Certificates) in \cite{dann2019policy}. We set $\rho = 0.2$ for our algorithm, which is the uncertain parameter used during the training. In Figure~\ref{fig:ARRLCvsORLC}, we plot the accumulated reward of both algorithms under different $p$ and perturbations. It can be seen that overall our ARRLC algorithm achieves a much higher reward than the ORLC algorithm. This demonstrates the robustness of our ARRLC algorithm to policy execution uncertainty.

We compare our algorithm with the robust TD algorithm in \cite{Klima2029robust}, which has no theoretical guarantee on sample complexity or regret. We set $\rho = 0.2$. In Figure~\ref{fig:ARRLCvsRobustTD}, we plot the accumulated reward of both algorithms under different $p$ and perturbations using a base-10 logarithmic scale on the x-axis and a linear scale on the y-axis. It can be seen that our ARRLC algorithm converges faster than the robust TD algorithm. This demonstrates the efficiency of our ARRLC algorithm to learn optimal policy under policy execution uncertainty.

\section{Conclusion}
In this paper, we have developed a novel approach for solving action robust RL problems with probabilistic policy execution uncertainty. We have theoretically proved the sample complexity bound and the regret bound of the algorithms. The upper bound of the sample complexity and the regret of proposed ARRLC algorithm match the lower bound up to logarithmic factors, which shows the minimax optimality of our algorithm. Moreover, we have carried out numerical experiments to validate our algorithm's robustness and efficiency, revealing that ARRLC surpasses non-robust algorithms and converges more rapidly than the robust TD algorithm when faced with action perturbations.

\bibliographystyle{unsrt}
{
\small
\bibliography{mybib}
}


\newpage
\appendix


\section{Proof of  Theorem~\ref{thm:Bellman}} \label{sec:ProofBellman}

The uncertainty set of the policy execution has the form in:
\begin{equation}
    \Pi^{\rho}(\pi) := \{\widetilde{\pi}| \forall s, \widetilde{\pi}_h(\cdot|s) = (1-\rho) \pi(\cdot|s) +\rho \pi'_h(\cdot|s), \pi'_h(\cdot|s) \in \Delta_{\mathcal{A}}  \}.
\end{equation}
We define \begin{eqnarray}C_h^{\pi,\pi',\rho}(s) &:=& \mathbb{E}\left[ \sum_{h'=h}^{H} R_{h'}(s_{h'},a_{h'}) | s_h=s , a_{h'} \sim \widetilde{\pi}_{h'}(\cdot|s_{h'})\right]\nonumber\\
D_h^{\pi,\pi',\rho}(s,a) &:=& \mathbb{E}\left[ \sum_{h'=h}^{H} R_{h'}(s_{h'},a_{h'}) | s_h=s ,a_h =a, a_{h'} \sim \widetilde{\pi}_{h'}(\cdot|s_{h'}) \right].\nonumber\end{eqnarray}

\paragraph{Robust Bellman Equation} First we prove the action robust Bellman equation holds for all policy $\pi$, state $s$ action $a$ and step $h$.
From the definition of the robust value function in \eqref{eq:robustV}, we have $V^{\pi}_{H+1}(s) = 0$, $\forall s \in \mathcal{S}$.

We prove the robust Bellman equation by building a policy ${\pi}^-$. Here, policy $\pi^-$ is the optimal adversarial policy towards the policy $\pi$. 

At step $H$, we set ${\pi}^-_H(s) = \argmin_{a \in \mathcal{A}} R_H(s,a)$. We have 
\begin{equation}
\begin{split}
V^{\pi}_{H}(s) &= \min_{\pi'} C_H^{\pi,\pi',\rho}(s) \\ &=  (1-\rho) [\mathbbm{D}_{\pi_H}R_{H}](s) +  \rho \min_{\pi'}  [\mathbbm{D}_{\pi'_H}R_{H}](s)\\
&=    (1-\rho) [\mathbbm{D}_{\pi_H}Q^{\pi}_{H}](s) + \rho\min_{a\in\mathcal{A}}Q^{\pi}_{H}(s,a)= C_H^{\pi,\pi^-,\rho}(s), 
\end{split}
\end{equation}
as $V_{H+1} = 0$. 

The robust Bellman equation holds at step $H$ and  $\min_{\pi'} \sum_{s} w(s) C_{H}^{\pi,\pi',\rho}(s) =  \sum_{s} w(s) \min_{\pi'} C_{H}^{\pi,\pi',\rho}(s) =  \sum_{s} w(s)  C_{H}^{\pi,\pi^-,\rho}(s)$ for any state $s$ and any weighted function $w: \mathcal{S} \rightarrow \Delta_{\mathcal{S}}$.

Suppose the robust Bellman equation holds at step $h+1$ and $\min_{\pi'} \sum_{s} w(s) C_{h+1}^{\pi,\pi',\rho}(s) =  \sum_{s} w(s) \min_{\pi'} C_{h+1}^{\pi,\pi',\rho}(s) =  \sum_{s} w(s)  C_{h+1}^{\pi,\pi^-,\rho}(s)$ for any state $s$ and any weighted function $w: \mathcal{S} \rightarrow \Delta_{\mathcal{S}}$.

Now we prove the robust Bellman equation holds at step $h$.
From the definition of the robust $Q$-function in \eqref{eq:robustQ} and the form of uncertainty set, we have 
\begin{equation} 
\begin{split}
        Q^{\pi}_{h}(s,a) = & \min_{\widetilde{\pi} \in \Pi(\pi)} \mathbb{E}\left[ \sum_{h'=h}^{H} R_{h'}(s_{h'},a_{h'}) | s_h=s , a_h = a, a_{h'} \sim \widetilde{\pi}_{h'}(\cdot|s_{h'})\right]\\
        =  & \min_{\pi'} D_h^{\pi,\pi',\rho}(s,a) \\
        = & R_{h}(s,a) +  \min_{\pi'} \mathbbm{E}_{s' \sim P_h(\cdot|s,a)} C_{h+1}^{\pi,\pi',\rho}(s) \\
        = & R_{h}(s,a) + \mathbbm{E}_{s' \sim P_h(\cdot|s,a)} \min_{\pi'}  C_{h+1}^{\pi,\pi',\rho}(s) \\
        = & R_{h}(s,a) + [{P}_{h}V_{h+1}^{\pi}](s,a).
\end{split}
\end{equation}
We also have that $ Q^{\pi}_{h}(s,a) = D_h^{\pi,\pi^-,\rho}(s,a) $.

Recall that a (stochastic) Markov policy is a set of $H$ maps $\pi := \{ \pi_{h}: \mathcal{S} \rightarrow \Delta_{\mathcal{A}} \}_{h \in [H]}$. 
From the definition of the robust value function in \eqref{eq:robustV} and the form of uncertainty set, we have 
\begin{equation}
\begin{split}
    V^{\pi}_{h}(s)  = & \min_{\widetilde{\pi} \in \Pi(\pi)} \mathbb{E}\left[ \sum_{h'=h}^{H} R_{h'}(s_{h'},a_{h'}) | s_h=s , a_{h'} \sim \widetilde{\pi}_{h'}(\cdot|s_{h'}) \right] \\
    = &  \min_{\pi'} C_h^{\pi,\pi',\rho}(s) \\
    = & \min_{{\pi'}_h } \min_{ \{ {\pi'}_{h'} \}_{h'=h+1}^H } C_h^{\pi,\pi',\rho}(s) \\
    \ge &  (1-\rho) \min_{ \{ {\pi'}_{h'} \}_{h'=h+1}^H} \mathbb{E}_{a \sim \pi_{h}(\cdot|s)} D_h^{\pi,\pi',\rho}(s,a) +  \rho \min_{{\pi'}_h } \min_{ \{ {\pi'}_{h'} \}_{h'=h+1}^H } \mathbb{E}_{a \sim \pi'_{h}(\cdot|s)} D_h^{\pi,\pi',\rho}(s,a)\\
    \ge &  (1-\rho) \mathbb{E}_{a \sim \pi_{h}(\cdot|s)} \min_{ \{ {\pi'}_{h'} \}_{h'=h+1}^H}  D_h^{\pi,\pi',\rho}(s,a) +  \rho \min_{{\pi'}_h }  \mathbb{E}_{a \sim \pi'_{h}(\cdot|s)} \min_{ \{ {\pi'}_{h'} \}_{h'=h+1}^H } D_h^{\pi,\pi',\rho}(s,a)\\
  =  & (1-\rho) [\mathbbm{D}_{\pi_h}Q^{\pi}_{h}](s) + \rho\min_{a\in\mathcal{A}}Q^{\pi}_{h}(s,a).
\end{split}
\end{equation}

We set ${\pi}^-_h(s) =\argmin_{a\in\mathcal{A}}Q_h^{\pi}(s,a) = \argmin_{a\in\mathcal{A}}D_h^{\pi,\pi^-,\rho}(s,a)$. 

At step $h$, we have 
\begin{equation}
\begin{split}
    V^{\pi}_{h}(s) \le  & C_h^{\pi,\pi^-,\rho}(s) \\
    = & (1-\rho) [\mathbbm{D}_{\pi_h}D_h^{\pi,\pi^-,\rho}](s) + \rho\min_{a\in\mathcal{A}}D_h^{\pi,\pi^-,\rho}(s,a) \\
    = & (1-\rho) [\mathbbm{D}_{\pi_h}Q^{\pi}_{h}](s) + \rho\min_{a\in\mathcal{A}}Q^{\pi}_{h}(s,a),
\end{split}
\end{equation}
where the last equation comes from the robust Bellman equation at step $h+1$ and  $$D_h^{\pi,\pi^-,\rho}(s,a) = R_{h}(s,a) + [{P}_{h}C_{h+1}^{\pi,\pi^-,\rho}](s,a) = R_{h}(s,a) + [{P}_{h}V_{h+1}^{\pi}](s,a).$$  Thus, the robust Bellman equation holds at step $h$.

Then, we prove the commutability of the expectation and the minimization operations at step $h$. For any weighted function $w$, we have $\min_{\pi'} \sum_{s} w(s) C_{h}^{\pi,\pi',\rho}(s) \ge \sum_{s} w(s) \min_{\pi'} C_{h}^{\pi,\pi',\rho}(s)$. Then,  $\min_{\pi'} \sum_{s} w(s) C_{h}^{\pi,\pi',\rho}(s) \le \sum_{s} w(s) C_{h}^{\pi,\pi^-,\rho}(s) = \sum_{s} w(s) \min_{\pi'} C_{h}^{\pi,\pi',\rho}(s)$.

By induction on $h = H, \cdots, 1$, we prove the robust Bellman equation.

\paragraph{Perfect Duality and Robust Bellman Optimality Equation} 
We now prove that the perfect duality holds and can be solved by the optimal robust Bellman equation.

The control problem in the LHS of \eqref{eq:duality} is equivalent to
\begin{equation}
\begin{split}
     & \max_{\pi}  \min_{\widetilde{\pi} \in \Pi^{\rho}(\pi)} \mathbb{E}\left[ \sum_{h'=h}^{H} R_{h'}(s_{h'},a_{h'}) | s_h=s , a_{h'} \sim \widetilde{\pi}_{h'}(\cdot|s_{h'}) \right]  =  \max_{\pi}  \min_{\pi'} C_h^{\pi,\pi',\rho}(s).
\end{split}    
\end{equation}
The control problem in the RHS of \eqref{eq:duality} is equivalent to
\begin{equation}
\begin{split}
     &  \min_{\widetilde{\pi} \in \Pi^{\rho}(\pi)} \max_{\pi}  \mathbb{E}\left[ \sum_{h'=h}^{H} R_{h'}(s_{h'},a_{h'}) | s_h=s , a_{h'} \sim \widetilde{\pi}_{h'}(\cdot|s_{h'}) \right]  =  \min_{\pi'} \max_{\pi}  C_h^{\pi,\pi',\rho}(s).
\end{split}    
\end{equation}

 For step $H$, we have
$ C_H^{\pi,\pi',\rho}(s) =  [\mathbbm{D}_{\left((1-\rho)\pi+\rho\pi'\right)_H}R_{H}](s) = (1-\rho)[\mathbbm{D}_{\pi_H} R_{H}](s) + \rho [\mathbbm{D}_{\pi'_H} R_{H}](s)$. Thus, we have
\begin{equation}
\begin{split}
      \max_{\pi}  \min_{\pi'}C_H^{\pi,\pi',\rho}(s) = & (1-\rho)\max_{\pi} [\mathbbm{D}_{\pi_H} R_{H}](s) + \rho \min_{\pi'} [\mathbbm{D}_{\pi'_H}R_{H}](s)\\
     = & (1-\rho)\max_{a \in \mathcal{A}} R_{H}(s,a) + \rho\min_{b \in \mathcal{A}} R_{H}(s,b),
\end{split}    
\end{equation}
and
\begin{equation}
\begin{split}
      \min_{\pi'} \max_{\pi} C_H^{\pi,\pi',\rho}(s) = & (1-\rho)\max_{\pi} [\mathbbm{D}_{\pi_H} R_{H}](s) + \rho \min_{\pi'} [\mathbbm{D}_{\pi'_H}R_{H}](s)\\
     = & (1-\rho)\max_{a \in \mathcal{A}} R_{H}(s,a) + \rho\min_{b \in \mathcal{A}} R_{H}(s,b).
\end{split}    
\end{equation}
At step $H$, the perfect duality holds for all $s$ and there always exists an optimal robust policy $\pi^*_H(s) = \argmax_{a \in \mathcal{A}} Q_H^*(s,a) = \argmax_{a \in \mathcal{A}} R_{H}(s,a)$ and its corresponding optimal adversarial policy $\pi^-_H(s) = \argmin_{a \in \mathcal{A}} R_{H}(s,a)$ which are deterministic. The action robust Bellman optimality equation holds at step $H$ for all state $s$ and action $a$. 

In addition, $\max_{\pi} \min_{\pi'} \sum_{s} w(s) C_{H}^{\pi,\pi',\rho}(s) =  \sum_{s} w(s) \max_{\pi} \min_{\pi'} C_{H}^{\pi,\pi',\rho}(s)$ for any weighted function $w: \mathcal{S} \rightarrow \Delta_{\mathcal{S}}$. This can be shown as
\begin{equation}
\begin{split}
     & \max_{\pi}  \min_{\pi'} \sum_{s \in \mathcal{S}} w(s) C_H^{\pi,\pi',\rho}(s) \\
     = & (1-\rho)\max_{\pi}  \sum_{s \in \mathcal{S}} w(s)[\mathbbm{D}_{\pi_H} R_{H}](s) + \rho \min_{\pi'}  \sum_{s \in \mathcal{S}} w(s) [\mathbbm{D}_{\pi'_H}R_{H}](s)\\
     = & (1-\rho)  \sum_{s \in \mathcal{S}} w(s) \max_{a \in \mathcal{A}} R_{H}(s,a) + \rho  \sum_{s \in \mathcal{S}} w(s) \min_{b \in \mathcal{A}} R_{H}(s,b).
\end{split}    
\end{equation}

Suppose that at steps from $h+1$ to $H$, the perfect duality holds for all $s$, the action robust Bellman optimality equation holds for all state $s$ and action $a$, there always exists an optimal robust policy $\pi^*_{h'} = \argmax_{a \in \mathcal{A}} Q_{h'}^*(s,a)$ and its corresponding optimal adversarial policy $\pi^-_{h'}(s) = \argmin_{a \in \mathcal{A}} Q_{h'}^*(s,a)$, $\forall h' \ge h+1$, which is deterministic , and $\max_{\pi} \min_{\pi'} \sum_{s} w(s) C_{h'}^{\pi,\pi',\rho}(s) =  \sum_{s} w(s) \max_{\pi} \min_{\pi'} C_{h'}^{\pi,\pi',\rho}(s)$ for any state $s$, any weighted function $w: \mathcal{S} \rightarrow \Delta_{\mathcal{S}}$ and any $h' \ge h+1$. We have $V_{h'}^*(s) = V_{h'}^{\pi^*}(s) = C_{h'}^{\pi^*,\pi^-,\rho}(s)$ and $Q_{h'}^*(s,a) = Q_{h'}^{\pi^*}(s,a) = D_{h'}^{\pi^*,\pi^-,\rho}(s,a)$ for any state $s$ and any $h' \ge h+1$.

We first prove that the robust Bellman optimality equation holds at step $h$. 

We have \begin{equation}\begin{split}Q_h^*(s,a) &= \max_{\pi} \min_{\pi'} D_h^{\pi,\pi',\rho}(s,a)\\ & =  \max_{\pi} \min_{\pi'} ( R_{h}(s,a) + [{P}_{h}C_{h+1}^{\pi,\pi',\rho}](s,a))\\ &= R_{h}(s,a) + [{P}_{h} (\max_{\pi} \min_{\pi'}C_{h+1}^{\pi,\pi',\rho})](s,a)\\ & = R_h(s,a) + [{P}_{h}V_{h+1}^{*}](s,a).\end{split}\end{equation}
and also $Q_{h}^*(s,a) = Q_{h}^{\pi^*}(s,a) = D_{h}^{\pi^*,\pi^-,\rho}(s,a)$.

From the robust Bellman equation, we have
\begin{equation}
\begin{split}
   \max_{\pi} V^{\pi}_{h}(s)  = & \max_{\pi} \left((1-\rho) [\mathbbm{D}_{\pi_h}Q^{\pi}_{h}](s) + \rho\min_{a\in\mathcal{A}}Q^{\pi}_{h}(s,a) \right) \\
      \le & (1-\rho) \max_{{\pi}_h } \max_{ \{ {\pi}_{h} \}_{h'=h+1}^H} [\mathbbm{D}_{\pi_h}Q^{\pi}_{h}](s)  + \rho \max_{ \{ {\pi}_{h} \}_{h'=h+1}^H} \min_{a\in\mathcal{A}}Q^{\pi}_{h}(s,a)  \\
      \le & (1-\rho) \max_{{\pi}_h } \max_{ \{ {\pi}_{h} \}_{h'=h+1}^H} [\mathbbm{D}_{\pi_h}Q^{\pi}_{h}](s)  + \rho \min_{a\in\mathcal{A}} \max_{ \{ {\pi}_{h} \}_{h'=h+1}^H} Q^{\pi}_{h}(s,a) \\
      \le & (1-\rho) \max_{{\pi}_h } [\mathbbm{D}_{\pi_h}Q^{*}_{h}](s)  + \rho \min_{a\in\mathcal{A}}  Q^{*}_{h}(s,a)  \\
      = & (1-\rho) \max_{a\in\mathcal{A}}  Q^{*}_{h}(s,a)  + \rho \min_{a\in\mathcal{A}}  Q^{*}_{h}(s,a).
\end{split}
\end{equation}
We set $\pi^*_h(s) = \max_{a\in\mathcal{A}}  Q^{*}_{h}(s,a)$. According to the robust bellman equation, we have
\begin{equation}
\begin{split}
 \max_{\pi} V^{\pi}_{h}(s) \ge V^{\pi^*}_{h}(s)  &=  (1-\rho) [\mathbbm{D}_{\pi^*_h} Q^{\pi^*}_{h}](s)  + \rho \min_{a\in\mathcal{A}}  Q^{\pi^*}_{h}(s,a) \\
 &=  (1-\rho) \max_{a\in\mathcal{A}}  Q^{\pi^*}_{h}(s,a)  + \rho \min_{a\in\mathcal{A}}  Q^{\pi^*}_{h}(s,a) \\
 & = (1-\rho) \max_{a\in\mathcal{A}}  Q^{*}_{h}(s,a)  + \rho \min_{a\in\mathcal{A}}  Q^{*}_{h}(s,a).
\end{split}
\end{equation}
Thus, the robust Bellman optimality equation holds at step $h$. There always exists an optimal robust policy $\pi^*_{h} = \argmax_{a \in \mathcal{A}} Q_{h}^*(s,a)$ and its corresponding optimal adversarial policy $\pi^-_{h}(s) = \argmin_{a \in \mathcal{A}} Q_{h}^*(s,a)$ which is deterministic so that $ C_h^{\pi^*,\pi^-,\rho}(s) = V_h^*(s)$.

Then, we prove the commutability of the expectation, the minimization and the maximization operations at step $h$. 

In the proof of robust Bellman equation, we have shown that $$ \min_{\pi'} \sum_{s} w(s) C_{h}^{\pi,\pi',\rho}(s) =  \sum_{s} w(s) \min_{\pi'} C_{h}^{\pi,\pi',\rho}(s)$$ for any policy $\pi$ and any weighted function $w$. Hence $$\max_{\pi} \min_{\pi'} \sum_{s} w(s) C_{h}^{\pi,\pi',\rho}(s) \sum_{s} = \max_{\pi} \sum_{s} w(s) \min_{\pi'} C_{h}^{\pi,\pi',\rho}(s).$$ 
First, we have $$\max_{\pi} \sum_{s} w(s) \min_{\pi'} C_{h}^{\pi,\pi',\rho}(s) \le  \sum_{s}  w(s) \max_{\pi} \min_{\pi'} C_{h}^{\pi,\pi',\rho}(s).$$ Then, we can show 
\begin{eqnarray}\max_{\pi} \sum_{s} w(s) \min_{\pi'} C_{h}^{\pi,\pi',\rho}(s) &\ge&  \sum_{s}  w(s) \min_{\pi'} C_{h}^{\pi^*,\pi',\rho}(s)\nonumber\\ &=& \sum_{s}  w(s) C_{h}^{\pi^*,\pi^-,\rho}(s) \nonumber\\ &=&  \sum_{s}  w(s) \max_{\pi} \min_{\pi'} C_{h}^{\pi,\pi',\rho}(s).\end{eqnarray} 

In summary, $$\max_{\pi} \min_{\pi'} \sum_{s} w(s) C_{h}^{\pi,\pi',\rho}(s) \sum_{s} = w(s) \max_{\pi} \min_{\pi'} C_{h}^{\pi,\pi',\rho}(s).$$

We can show the perfect duality at step $h$ by
\begin{equation}
\begin{split}
       \max_{\pi}  \min_{\pi'} C_h^{\pi,\pi',\rho}(s) =  C_h^{\pi^*,\pi^-,\rho}(s) = \max_{\pi} C_h^{\pi,\pi^-,\rho}(s) \ge \min_{\pi'} \max_{\pi}  C_h^{\pi,\pi',\rho}(s).
\end{split}
\end{equation}

 By induction on $h = H, \cdots, 1$, we prove Theorem~\ref{thm:Bellman}.

\section{Proof for Action Robust Reinforcement Learning with Certificates}

In this section, we prove Theorem~\ref{thm:ARRLC}. Recall that we use 
$\overline{Q}_h^k$,$\overline{V}_h^k$,$\underline{Q}_h^k$,$\underline{V}_h^k$, $N_h^k$, $\hat{P}_h^k$,$\hat{r}_h^k$ and $\theta_h^k$ to denote the values of $\overline{Q}_h$,$\overline{V}_h$,$\underline{Q}_h$,$\underline{V}_h$, $\max\{N_h,1\}$, $\hat{P}_h$, ${r}_h$ and $\theta_h$ at the beginning of the $k$-th episode in Algorithm~\ref{alg:ARRLC}.

\subsection{Proof of monotonicity}

\subsubsection{Proof of Lemma~\ref{lem:event_mb}}
When $N^k_h(s, a) \le 1$, \eqref{eq:eventPV1}, \eqref{eq:eventPV2} and \eqref{eq:eventR} hold trivially by the bound of the rewards and value functions.

For every $h \in [H]$ the empiric Bernstein inequality combined with a union bound argument, to take into account that $N^k_h(s, a) > 1$ is a random number, leads to the following inequality w.p. $1 - SAH\delta$ (see Theorem 4 in \cite{maurer2009empirical})
\begin{equation}
\begin{split}
 \left| (\hat{P}_h^k - P_h)V^*_{h+1}(s,a)  \right| \le \sqrt{\frac{2\mathbbm{V}_{\hat{P}_h^k}V_{h+1}^*(s,a) \iota}{N_h^k(s,a)  }} + \frac{7H \iota}{3(N_h^k(s,a) )},
\end{split}
\end{equation}
and
\begin{equation}
\begin{split}
 \left| (\hat{P}_h^k - P_h)V^{\overline{\pi}^k}_{h+1}(s,a)  \right| \le \sqrt{\frac{2\mathbbm{V}_{\hat{P}_h^k}V_{h+1}^{\overline{\pi}^k}(s,a) \iota}{N_h^k(s,a)  }} + \frac{7H \iota}{3(N_h^k(s,a) )}.
\end{split}
\end{equation}
Similarly, with Azuma's inequality, w.p. $1 - SAH\delta$
\begin{equation} 
     \left| \hat{r}_h^k(s,a) - R_h(s,a)  \right| \le  \sqrt{\frac{2 Var(r_h^k(s,a)) \iota}{N_h^k(s,a)  }} + \frac{7 \iota}{3(N_h^k(s,a) )} 
     \le  \sqrt{\frac{2 \hat{r}_h^k(s,a) \iota}{N_h^k(s,a)  }} + \frac{7 \iota}{3(N_h^k(s,a) )},
\end{equation}
where $Var(r_h^k(s,a))$  is the empirical variance of $R_h(s, a)$ computed by the $N_h^k (s, a)$ samples and $Var(r_h^k(s,a)) \le \hat{r}_h^k(s,a)$ . 

\subsubsection{Proof of Lemma~\ref{lem:Monot_mb}} \label{sec:proof_monot_mb}
We first prove that $\overline{Q}_h^k (s,a) \ge Q_h^*(s,a)$ for all $(s, a, h, k) \in S \times A \times [H] \times[K]$, by
backward induction conditioned on the event $\mathcal{E}^R \cap \mathcal{E}^{PV}$.  Firstly, the conclusion holds for $h = H + 1$ because $\overline{V}_{H+1}(s)=\underline{V}_{H+1}(s) =0 $  and $\overline{Q}_{H+1}(s,a)=\underline{Q}_{H+1}(s,a) = 0$ for all $s$ and $a$. For $h \in [H]$, assuming the conclusion holds for $h + 1$, by Algorithm~\ref{alg:ARRLC}, we have
\begin{equation} \label{eq:mbQ*}
\begin{split}
   & \hat{r}_h^k(s,a) + \hat{P}_h^k\overline{V}_{h+1}(s,a) + \theta_h^k(s,a) - Q_h^*(s,a) \\
  = & \hat{r}_h^k(s,a) + \hat{P}_h^k\overline{V}_{h+1}(s,a) + \theta_h^k(s,a) - R_h(s,a) - P_h V_{h+1}^*(s,a) \\
  = & \hat{r}_h^k(s,a) - R_h(s,a) + \hat{P}_h^k\left(\overline{V}_{h+1}-V_{h+1}^*\right)(s,a)   + (\hat{P}_h^k - P_h) V_{h+1}^*(s,a) + \theta_h^k(s,a) \\
  \ge & (\hat{P}_h^k - P_h) V_{h+1}^*(s,a) + \sqrt{\frac{2\mathbbm{V}_{\hat{P}_h^k}[(\overline{V}^k_{h+1}+\underline{V}^k_{h+1})/2](s,a)\iota }{N_h^k(s,a)  }}  + \frac{  \hat{P}_h^k \left(\overline{V}^k_{h+1}-\underline{V}^k_{h+1} \right)(s,a)}{H} + \frac{ 8H^2 \iota}{N_h^k(s,a)  } \\
  \ge & \sqrt{\frac{2\mathbbm{V}_{\hat{P}_h^k}[(\overline{V}^k_{h+1}+\underline{V}^k_{h+1})/2](s,a)\iota }{N_h^k(s,a)  }} + \frac{  \hat{P}_h^k \left(\overline{V}^k_{h+1}-\underline{V}^k_{h+1} \right)(s,a)}{H} + \frac{ 8H^2 \iota}{N_h^k(s,a)  } - \sqrt{\frac{2\mathbbm{V}_{\hat{P}_h^k}V_{h+1}^*(s,a) \iota}{N_h^k(s,a)  }},
\end{split}
\end{equation}
where the first inequality comes from event $\mathcal{E}^R$, $\overline{V}_{h+1}(s) \ge V_{h+1}^*(s)$ and the definition of $\theta_h^k(s,a)$ and the last inequality from event $\mathcal{E}^{PV}$.
By the relation of $V$-values in the step $(h + 1)$,
\begin{equation}
\begin{split}
     & \left|\mathbbm{V}_{\hat{P}_h^k}\left(\frac{\overline{V}^k_{h+1}+\underline{V}^k_{h+1}}{2}\right)(s,a)  - \mathbbm{V}_{\hat{P}_h^k}V_{h+1}^*(s,a) \right|\\
     \le &\left| [\hat{P}_h^k(\overline{V}^k_{h+1}+\underline{V}^k_{h+1})/2]^2 - (\hat{P}_h^kV_{h+1}^*)^2 \right|(s,a)  + \left| \hat{P}_h^k[(\overline{V}^k_{h+1}+\underline{V}^k_{h+1})/2]^2 - \hat{P}_h^k(V_{h+1}^*)^2 \right|(s,a) \\
     \le & 4H \hat{P}_h^k \left|(\overline{V}^k_{h+1}+\underline{V}^k_{h+1})/2 -V_{h+1}^*  \right|(s,a) \\
     \le & 2H \hat{P}_h^k \left(\overline{V}^k_{h+1}-\underline{V}^k_{h+1} \right)(s,a) 
\end{split}
\end{equation}
and 
\begin{equation} \label{eq:Vardev}
\begin{split}
     &  \sqrt{\frac{2\mathbbm{V}_{\hat{P}_h^k}V_{h+1}^*(s,a) \iota}{N_h^k(s,a)  }} \\
     \le & \sqrt{\frac{2\mathbbm{V}_{\hat{P}_h^k}[(\overline{V}^k_{h+1}+\underline{V}^k_{h+1})/2](s,a)\iota + 4H \hat{P}_h^k \left(\overline{V}^k_{h+1}-\underline{V}^k_{h+1} \right)(s,a) \iota}{N_h^k(s,a)  }} \\
     \le & \sqrt{\frac{2\mathbbm{V}_{\hat{P}_h^k}[(\overline{V}^k_{h+1}+\underline{V}^k_{h+1})/2](s,a)\iota }{N_h^k(s,a)  }} + \sqrt{\frac{ 4H \hat{P}_h^k \left(\overline{V}^k_{h+1}-\underline{V}^k_{h+1} \right)(s,a) \iota}{N_h^k(s,a)  }} \\
      \le & \sqrt{\frac{2\mathbbm{V}_{\hat{P}_h^k}[(\overline{V}^k_{h+1}+\underline{V}^k_{h+1})/2](s,a)\iota }{N_h^k(s,a)  }} + \frac{  \hat{P}_h^k \left(\overline{V}^k_{h+1}-\underline{V}^k_{h+1} \right)(s,a)}{H} + \frac{ 8H^2 \iota}{N_h^k(s,a)  }.
\end{split}
\end{equation}
Plugging \eqref{eq:Vardev} back into \eqref{eq:mbQ*}, we have $\hat{r}_h^k(s,a) + \hat{P}_h^k\overline{V}_{h+1}(s,a) + \theta_h^k(s,a) \ge Q_h^*(s,a)$. Thus, $\overline{Q}_h^k(s,a) = \min\{H-h+1, \hat{r}_h^k(s,a) + \hat{P}_h^k\overline{V}_{h+1}^k(s,a) + \theta_h^k(s,a) \} \ge Q_h^*(s,a)$. 

From the definition of $\overline{V}_{h}^k(s)$ and $\overline{\pi}_h^k$, we have 
\begin{equation}
    \begin{split}
      \overline{V}_{h}^k(s) = &  (1-\rho)\overline{Q}_h^k(s, \overline{\pi}_h^k(s)) + \rho \overline{Q}_h^k(s, \underline{\pi}_h^k(s)) \\
      \ge & (1-\rho)\overline{Q}_h^k(s, {\pi}_h^*(s)) + \rho {Q}_h^*(s, \underline{\pi}_h^k(s)) \\
      \ge & (1-\rho){Q}_h^*(s, {\pi}_h^*(s)) + \rho \min_{a \in \mathcal{A}} {Q}_h^*(s, a) = {V}_h^*(s).
    \end{split}
\end{equation}

Similarly, we can prove that $\underline{Q}_h^k (s,a) \le Q_h^{\overline{\pi}^k}(s,a)$ and $\underline{V}_h^k (s) \le V_h^{\overline{\pi}^k}(s)$.

\begin{equation} \label{eq:mbQk}
\begin{split}
   & \hat{r}_h^k(s,a) + \hat{P}_h^k\underline{V}_{h+1}(s,a) - \theta_h^k(s,a) - Q_h^{\overline{\pi}^k}(s,a) \\
  = & \hat{r}_h^k(s,a) + \hat{P}_h^k\underline{V}_{h+1}(s,a) - \theta_h^k(s,a) - R_h(s,a) - P_h V_{h+1}^{\overline{\pi}^k}(s,a) \\
  = & \hat{r}_h^k(s,a) - R_h(s,a) + \hat{P}_h^k\left(\underline{V}_{h+1}-V_{h+1}^{\overline{\pi}^k}\right)(s,a)   + (\hat{P}_h^k - P_h) V_{h+1}^{\overline{\pi}^k}(s,a) - \theta_h^k(s,a) \\
  \le & (\hat{P}_h^k - P_h) V_{h+1}^{\overline{\pi}^k}(s,a) - \sqrt{\frac{2\mathbbm{V}_{\hat{P}_h^k}[(\overline{V}^k_{h+1}+\underline{V}^k_{h+1})/2](s,a)\iota }{N_h^k(s,a)  }} \\
  & - \frac{  \hat{P}_h^k \left(\overline{V}^k_{h+1}-\underline{V}^k_{h+1} \right)(s,a)}{H} - \frac{ 8H^2 \iota}{N_h^k(s,a)  } \\
  \le &\sqrt{\frac{2\mathbbm{V}_{\hat{P}_h^k}V_{h+1}^{\overline{\pi}^k}(s,a) \iota}{N_h^k(s,a)  }} - \sqrt{\frac{2\mathbbm{V}_{\hat{P}_h^k}[(\overline{V}^k_{h+1}+\underline{V}^k_{h+1})/2](s,a)\iota }{N_h^k(s,a)  }} \\
  &- \frac{  \hat{P}_h^k \left(\overline{V}^k_{h+1}-\underline{V}^k_{h+1} \right)(s,a)}{H} - \frac{ 8H^2 \iota}{N_h^k(s,a)  }   \le  0,
\end{split}
\end{equation}
and
\begin{equation}
    \begin{split}
      \underline{V}_{h}^k(s) = &  (1-\rho)\underline{Q}_h^k(s, \overline{\pi}_h^k(s)) + \rho \underline{Q}_h^k(s, \underline{\pi}_h^k(s)) \\
      \le & (1-\rho){Q}_h^{\overline{\pi}^k}(s, \overline{\pi}_h^k(s)) +  \rho \min_{a \in \mathcal{A}} \underline{Q}_h^k(s, a) \\
      \le & (1-\rho){Q}_h^{\overline{\pi}^k}(s, \overline{\pi}_h^k(s)) + \rho \underline{Q}_h^k(s, \argmin_{a \in \mathcal{A}}{Q}_h^{\overline{\pi}^k}(s, a) ) \\
      \le & (1-\rho){Q}_h^{\overline{\pi}^k}(s, \overline{\pi}_h^k(s)) + \rho \min_{a \in \mathcal{A}} {Q}_h^{\overline{\pi}^k}(s, a) = {V}_h^{\overline{\pi}^k}(s).
    \end{split}
\end{equation}

\subsection{Regret Analysis}
\subsubsection{Proof of Lemma~\ref{lem:regdec}}
We consider the event $\mathcal{E}^R \cap \mathcal{E}^{PV}$. The following analysis will be done assuming the successful event $\mathcal{E}^R \cap \mathcal{E}^{PV}$ holds.
By Lemma~\ref{lem:Monot_mb}, the regret can be bounded by $ Regret(K):= \sum_{k=1}^K ( V_{1}^*(s_1^k) -  {V}_1^{\overline{\pi}^k}(s_1^k))\le  \sum_{k=1}^K (\overline{V}_{1}^k(s_1^k) -  \underline{V}_1^{k}(s_1^k)) $.

By the update steps in Algorithm~\ref{alg:ARRLC}, we have 
\begin{equation} \label{eq:regdecmboverh}
    \begin{split}
& \overline{V}_{h}^k(s_h^k) -  \underline{V}_h^{k}(s_h^k) \\
= &  (1-\rho)\overline{Q}_h^k(s_h^k, \overline{\pi}_h^k(s_h^k)) + \rho \overline{Q}_h^k(s_h^k, \underline{\pi}_h^k(s_h^k)) - (1-\rho)\underline{Q}_h^k(s_h^k, \overline{\pi}_h^k(s_h^k)) - \rho \underline{Q}_h^k(s_h^k, \underline{\pi}_h^k(s_h^k)) \\
\le & [ \mathbbm{D}_{\widetilde{\pi}_h^k}\hat{P}_h^k(\overline{V}_{h+1}^k - \underline{V}_{h+1}^k)](s_h^k)  + 2\mathbbm{D}_{\widetilde{\pi}_h^k} \theta_h(s_h^k) \\
= & [ \mathbbm{D}_{\widetilde{\pi}_h^k}\hat{P}_h^k(\overline{V}_{h+1}^k - \underline{V}_{h+1}^k)](s_h^k) - [\hat{P}_h^k(\overline{V}_{h+1}^k - \underline{V}_{h+1}^k)](s_h^k, a_h^k) + 2\mathbbm{D}_{\widetilde{\pi}_h^k} \theta_h(s_h^k)\\
 &+ [\hat{P}_h^k(\overline{V}_{h+1}^k - \underline{V}_{h+1}^k)](s_h^k, a_h^k) \\
=& [ \mathbbm{D}_{\widetilde{\pi}_h^k}\hat{P}_h^k(\overline{V}_{h+1}^k - \underline{V}_{h+1}^k)](s_h^k) - [\hat{P}_h^k(\overline{V}_{h+1}^k - \underline{V}_{h+1}^k)](s_h^k, a_h^k) + 2\mathbbm{D}_{\widetilde{\pi}_h^k} \theta_h(s_h^k)\\
 &+ [\hat{P}_h^k(\overline{V}_{h+1}^k - \underline{V}_{h+1}^k)](s_h^k, a_h^k) - c_1{P}_h(\overline{V}_{h+1}^k -  \underline{V}_{h+1}^k)(s_h^k,a_h^k) \\
& + c_1 {P}_h(\overline{V}_{h+1}^k - \underline{V}_{h+1}^k)(s_h^k,a_h^k) -  c_2(\overline{V}_{h+1}^k - \underline{V}_{h+1}^k)(s_{h+1}^k) + c_2 (\overline{V}_{h+1}^k - \underline{V}_{h+1}^k)(s_{h+1}^k) \\
=& [ \mathbbm{D}_{\widetilde{\pi}_h^k}\hat{P}_h^k(\overline{V}_{h+1}^k - \underline{V}_{h+1}^k)](s_h^k) - [\hat{P}_h^k(\overline{V}_{h+1}^k - \underline{V}_{h+1}^k)](s_h^k, a_h^k) \\
 &+ [\hat{P}_h^k(\overline{V}_{h+1}^k - \underline{V}_{h+1}^k)](s_h^k, a_h^k) - c_1{P}_h(\overline{V}_{h+1}^k -  \underline{V}_{h+1}^k)(s_h^k,a_h^k) \\
& + c_1 {P}_h(\overline{V}_{h+1}^k - \underline{V}_{h+1}^k)(s_h^k,a_h^k) -  c_2(\overline{V}_{h+1}^k - \underline{V}_{h+1}^k)(s_{h+1}^k) + c_2 (\overline{V}_{h+1}^k - \underline{V}_{h+1}^k)(s_{h+1}^k) \\
&+ 2(1-\rho)\sqrt{\frac{2\mathbbm{V}_{\hat{P}_h^k}[(\overline{V}^k_{h+1}+\underline{V}^k_{h+1})/2](s_h^k,\overline{\pi}_h^k(s_h^k))\iota }{N_h^k(s_h^k,\overline{\pi}_h^k(s_h^k))  }} + 2(1-\rho) \sqrt{\frac{2 \hat{r}_h^k(s_h^k,\overline{\pi}_h^k(s_h^k)) \iota}{N_h^k(s_h^k,\overline{\pi}_h^k(s_h^k))  }}  \\
& +(1-\rho)\hat{P}_h^k(\overline{V}_{h+1}^k - \underline{V}_{h+1}^k)(s_h^k,\overline{\pi}_h^k(s_h^k))/H  + \frac{2(1-\rho)(24H^2+7H+7)\iota}{3N_h^k(s_h^k,\overline{\pi}_h^k(s_h^k)) )} \\
& + 2\rho\sqrt{\frac{2\mathbbm{V}_{\hat{P}_h^k}[(\overline{V}^k_{h+1}+\underline{V}^k_{h+1})/2](s_h^k,\underline{\pi}_h^k(s_h^k))\iota }{N_h^k(s_h^k,\underline{\pi}_h^k(s_h^k))  }} + 2\rho\sqrt{\frac{2 \hat{r}_h^k(s_h^k,\underline{\pi}_h^k(s_h^k)) \iota}{N_h^k(s_h^k,\underline{\pi}_h^k(s_h^k))  }}\\
& + \rho\hat{P}_h^k(\overline{V}_{h+1}^k - \underline{V}_{h+1}^k)(s_h^k,\underline{\pi}_h^k(s_h^k))/H  + \frac{2\rho(24H^2+7H+7)\iota}{3N_h^k(s_h^k,\underline{\pi}_h^k(s_h^k)) )} \\
= & (1+1/H)[ \mathbbm{D}_{\widetilde{\pi}_h^k}\hat{P}_h^k(\overline{V}_{h+1}^k - \underline{V}_{h+1}^k)](s_h^k) - (1+1/H) [\hat{P}_h^k(\overline{V}_{h+1}^k - \underline{V}_{h+1}^k)](s_h^k,a_h^k) \\
 &+ \underbrace{ (1+1/H) [\hat{P}_h^k(\overline{V}_{h+1}^k - \underline{V}_{h+1}^k)](s_h^k, a_h^k) - c_1{P}_h(\overline{V}_{h+1}^k -  \underline{V}_{h+1}^k)(s_h^k,a_h^k) }_{(a)}\\
& + c_1 {P}_h(\overline{V}_{h+1}^k - \underline{V}_{h+1}^k)(s_h^k,a_h^k) -  c_2(\overline{V}_{h+1}^k - \underline{V}_{h+1}^k)(s_{h+1}^k) + c_2 (\overline{V}_{h+1}^k - \underline{V}_{h+1}^k)(s_{h+1}^k) \\
&+ 2(1-\rho) \underbrace{\sqrt{\frac{2\mathbbm{V}_{\hat{P}_h^k}[(\overline{V}^k_{h+1}+\underline{V}^k_{h+1})/2](s_h^k,\overline{\pi}_h^k(s_h^k))\iota }{N_h^k(s_h^k,\overline{\pi}_h^k(s_h^k))  }} }_{(b1)} + 2(1-\rho) \sqrt{\frac{2 \hat{r}_h^k(s_h^k,\overline{\pi}_h^k(s_h^k)) \iota}{N_h^k(s_h^k,\overline{\pi}_h^k(s_h^k))  }}  \\
& + \frac{2(1-\rho)(24H^2+7H+7)\iota}{3N_h^k(s_h^k,\overline{\pi}_h^k(s_h^k)) )} + 2\rho \underbrace{\sqrt{\frac{2\mathbbm{V}_{\hat{P}_h^k}[(\overline{V}^k_{h+1}+\underline{V}^k_{h+1})/2](s_h^k,\underline{\pi}_h^k(s_h^k))\iota }{N_h^k(s_h^k,\underline{\pi}_h^k(s_h^k))  }} }_{(b2)}\\
& + 2\rho\sqrt{\frac{2 \hat{r}_h^k(s_h^k,\underline{\pi}_h^k(s_h^k)) \iota}{N_h^k(s_h^k,\underline{\pi}_h^k(s_h^k))  }} +  \frac{2\rho(24H^2+7H+7)\iota}{3N_h^k(s_h^k,\underline{\pi}_h^k(s_h^k)) )} .
    \end{split}
\end{equation}

\paragraph{Bound of the error of the empirical probability estimator (a)}
By Bennett’s inequality, we have that w.p. $1 - S\delta$
\begin{equation} \label{eq:bennet_mb}
    |\hat{P}_h^k(s'|s,a) - P_h(s'|s,a)| \le \sqrt{\frac{2P_h(s'|s,a) \iota}{N_h^k(s,a)}} +\frac{\iota}{3N_h^k(s,a)}
\end{equation}
holds for all $s,a,h,k,s'$.

Thus, we have that
\begin{equation} \label{eq:mbeepe}
    \begin{split}
     & (\hat{P}_h^k - P_h)(\overline{V}_{h+1}^k - \underline{V}_{h+1}^k)(s,a) \\
      =& \sum_{s'} ( \hat{P}_h^k(s'|s,a) - P_h(s'|s,a) )(\overline{V}_{h+1}^k(s') - \underline{V}_{h+1}^k(s') ) \\
      \le & \sum_{s'} \sqrt{\frac{2P_h(s'|s,a) \iota}{N_h^k(s,a)}} (\overline{V}_{h+1}^k(s') - \underline{V}_{h+1}^k(s')) +\frac{SH\iota}{3N_h^k(s,a)} \\
    \le & \sum_{s'}\left(\frac{P_h(s'|s,a) \iota}{H} + \frac{H}{2N_h^k(s,a)} \right) \left(\overline{V}_{h+1}^k(s') - \underline{V}_{h+1}^k(s')\right) +\frac{SH\iota}{3N_h^k(s,a)} \\
    \le & P_h( \overline{V}_{h+1}^k - \underline{V}_{h+1}^k)(s,a)/H +\frac{SH^2}{2N_h^k(s,a)}  +\frac{SH\iota}{3N_h^k(s,a)} \\
    \le & P_h( \overline{V}_{h+1}^k - \underline{V}_{h+1}^k)(s,a)/H +\frac{SH^2\iota}{N_h^k(s,a)} ,
    \end{split}
\end{equation}
where the second inequality is due to AM-GM inequality.

\paragraph{Bound of the error of the empirical variance estimator (b1) \& (b2)} Here, we bound $\mathbbm{V}_{\hat{P}_h^k}[(\overline{V}^k_{h+1}+\underline{V}^k_{h+1})/2](s_h^k,a_h^k)$. 

Recall that $C_h^{\pi,\pi',\rho}(s) = \mathbb{E}\left[ \sum_{h'=h}^{H} R_{h'}(s_{h'},a_{h'}) | s_h=s , a_{h'} \sim \widetilde{\pi}_{h'}(\cdot|s_{h'}) \right]$  in Appendix~\ref{sec:ProofBellman}. Set ${\pi}^{k*}$ here is the optimal policy towards the adversary policy $\underline{\pi}^k$ with ${\pi}^{k*}_{h}(s) = \argmax_\pi {C}^{{\pi},\underline{\pi}^k,\rho}_{h}(s)$. Similar to the proof in Appendix~\ref{sec:proof_monot_mb}, we can show that $\overline{V}_{h}^k(s) \ge {C}^{{\pi}^{k*},\underline{\pi}^k,\rho}_{h}(s)$.  We also have that ${C}^{{\pi}^{k*},\underline{\pi}^k,\rho}_{h}(s) = \max_{\pi} {C}^{{\pi},\underline{\pi}^k,\rho}_{h}(s) \ge {C}^{\overline{\pi}^k,\underline{\pi}^k,\rho}_{h}(s) \ge V_h^{\overline{\pi}^k}(s) \ge \underline{V}_h^{k}(s)$ . For any $(s,a,h,k) \in \mathcal{S} \times \mathcal{A} \times [H] \times [K]$, under event $\mathcal{E}^R \cap \mathcal{E}^{PV}$, 
\begin{equation}\label{eq:bzd2}
    \begin{split}
     & \mathbbm{V}_{\hat{P}_h^k}[(\overline{V}^k_{h+1}+\underline{V}^k_{h+1})/2](s,a) - \mathbbm{V}_{{P}_h}{C}^{{\pi}^{k*},\underline{\pi}^k,\rho}_{h+1}(s,a) \\
     =& \hat{P}_h^k[(\overline{V}^k_{h+1}+\underline{V}^k_{h+1})/2]^2(s,a) - [\hat{P}_h^k(\overline{V}^k_{h+1}+\underline{V}^k_{h+1})/2]^2(s,a) \\
     & - {P}_h({C}^{{\pi}^{k*},\underline{\pi}^k,\rho}_{h+1})^2(s,a) + ({P}_h{C}^{{\pi}^{k*},\underline{\pi}^k,\rho}_{h+1})^2(s,a)\\
      \le & [ \hat{P}_h^k(\overline{V}^k_{h+1})^2 - (\hat{P}_h^k\underline{V}^k_{h+1})^2 - P_h(\underline{V}^k_{h+1})^2 + (P_h\overline{V}^k_{h+1})^2] (s,a) \\
      \le & |  (\hat{P}_h^k-P_h)(\overline{V}^k_{h+1})^2|(s,a) + |(P_h\underline{V}^k_{h+1})^2 - (\hat{P}_h^k\underline{V}^k_{h+1})^2 |(s,a)\\
      & + P_h|(\overline{V}^k_{h+1})^2 -(\underline{V}^k_{h+1})^2|(s,a) + |(P_h\overline{V}^k_{h+1})^2 - (P_h\underline{V}^k_{h+1})^2 |(s,a),
    \end{split}
\end{equation}

where the first inequality is due $\overline{V}_h^{k}(s) \ge {C}^{{\pi}^{k*},\underline{\pi}^k,\rho}_{h}(s) \ge \underline{V}_h^{k}(s)$.
The result of \cite{weissman2003inequalities} combined with a union bound on $N^k_h (s, a) \in [K ]$ implies w.p $1 - \delta$
\begin{equation} \label{eq:weissman_mb}
    \lVert \hat{P}_h^k(\cdot|s,a) - P_h(\cdot|s,a) \rVert_{1} \le \sqrt{\frac{2 S \iota}{N_h^k(s,a)}}
\end{equation}
holds for all $s,a,h,k$. 

These terms can be bounded separately by 
\begin{equation}
\begin{split}
  & |  (\hat{P}_h^k-P_h)(\overline{V}^k_{h+1})^2|(s,a) \le H^2\sqrt{\frac{2S\iota}{N_h^k(s,a)}},  \\
    & |(P_h\underline{V}^k_{h+1})^2 - (\hat{P}_h^k\underline{V}^k_{h+1})^2 |(s,a) \le 2H|(P_h-\hat{P}_h^k)\underline{V}^k_{h+1} | \le  2H^2 \sqrt{\frac{2S\iota}{N_h^k(s,a)}},  \\
   &  P_h|(\overline{V}^k_{h+1})^2 -(\underline{V}^k_{h+1})^2|(s,a) \le 2H P_h(\overline{V}^k_{h+1} - \underline{V}^k_{h+1})(s,a),\\
   & |(P_h\overline{V}^k_{h+1})^2 - (P_h\underline{V}^k_{h+1})^2 |(s,a) \le 2H P_h(\overline{V}^k_{h+1} - \underline{V}^k_{h+1})(s,a),
\end{split}
\end{equation}
where the first two inequality is due to \eqref{eq:weissman_mb}.
In addition, $3H^2 \sqrt{\frac{2S\iota}{N_h^k(s,a)}}\le 1 + \frac{9 SH^4 \iota}{2N_h^k(s,a)}$.
Thus, we have
\begin{equation} \label{eq:mbeeve}
\begin{split}
   & (1-\rho)\sqrt{\frac{\mathbbm{V}_{\hat{P}_h^k}[(\overline{V}^k_{h+1}+\underline{V}^k_{h+1})/2](s_h^k,\overline{\pi}_h^k(s_h^k) )\iota }{N_h^k(s_h^k,\overline{\pi}_h^k(s_h^k))  }} + \rho\sqrt{\frac{\mathbbm{V}_{\hat{P}_h^k}[(\overline{V}^k_{h+1}+\underline{V}^k_{h+1})/2](s_h^k,\underline{\pi}_h^k(s_h^k))\iota }{N_h^k(s_h^k,\underline{\pi}_h^k(s_h^k))  }}\\
\le &  (1-\rho) \sqrt{\frac{\mathbbm{V}_{{P}_h}{C}^{{\pi}^{k*},\underline{\pi}^k,\rho}_{h+1}(s_h^k,\overline{\pi}_h^k(s_h^k))\iota }{N_h^k(s_h^k, \overline{\pi}_h^k(s_h^k))  }} + \rho \sqrt{\frac{\mathbbm{V}_{{P}_h}{C}^{{\pi}^{k*},\underline{\pi}^k,\rho}_{h+1}(s_h^k,\underline{\pi}_h^k(s_h^k))\iota }{N_h^k(s_h^k, \underline{\pi}_h^k(s_h^k))  }} \\
& + (1-\rho) \sqrt{\frac{4H P_h(\overline{V}^k_{h+1} - \underline{V}^k_{h+1})(s_h^k,\overline{\pi}_h^k(s_h^k)) \iota }{N_h^k(s_h^k,\overline{\pi}_h^k(s_h^k))  }}+  \rho \sqrt{\frac{4H P_h(\overline{V}^k_{h+1} - \underline{V}^k_{h+1})(s_h^k,\underline{\pi}_h^k(s_h^k)) \iota }{N_h^k(s_h^k,\underline{\pi}_h^k(s_h^k))  }} \\
&+ (1-\rho)\sqrt{\frac{1}{N_h^k(s_h^k, \overline{\pi}_h^k(s_h^k))  } }  +\rho\sqrt{\frac{1}{N_h^k(s_h^k, \underline{\pi}_h^k(s_h^k))  } }   + \frac{(1-\rho) \sqrt{9SH^4\iota/2}}{N_h^k(s_h^k,\overline{\pi}_h^k(s_h^k))  }  + \frac{\rho \sqrt{9SH^4\iota/2}}{N_h^k(s_h^k,\underline{\pi}_h^k(s_h^k))  }  \\
\le &  (1-\rho) \sqrt{\frac{\mathbbm{V}_{{P}_h}{C}^{{\pi}^{k*},\underline{\pi}^k,\rho}_{h+1}(s_h^k,\overline{\pi}_h^k(s_h^k))\iota }{N_h^k(s_h^k, \overline{\pi}_h^k(s_h^k))  }} + \rho \sqrt{\frac{\mathbbm{V}_{{P}_h}{C}^{{\pi}^{k*},\underline{\pi}^k,\rho}_{h+1}(s_h^k,\underline{\pi}_h^k(s_h^k))\iota }{N_h^k(s_h^k, \underline{\pi}_h^k(s_h^k))  }} \\
& + (1-\rho) \left(\frac{P_h(\overline{V}^k_{h+1} - \underline{V}^k_{h+1})(s_h^k,\overline{\pi}_h^k(s_h^k)) }{2\sqrt{2}H} + \frac{2\sqrt{2}H^2\iota}{N_h^k(s_h^k,\overline{\pi}_h^k(s_h^k))  } \right) \\
& + \rho \left(\frac{P_h(\overline{V}^k_{h+1} - \underline{V}^k_{h+1})(s_h^k,\underline{\pi}_h^k(s_h^k)) }{2\sqrt{2}H} + \frac{2\sqrt{2}H^2\iota}{N_h^k(s_h^k,\underline{\pi}_h^k(s_h^k))  } \right) \\
&+ (1-\rho)\sqrt{\frac{1}{N_h^k(s_h^k, \overline{\pi}_h^k(s_h^k))  } }  +\rho\sqrt{\frac{1}{N_h^k(s_h^k, \underline{\pi}_h^k(s_h^k))  } }  \\
& + \frac{(1-\rho) \sqrt{9SH^4\iota/2}}{N_h^k(s_h^k,\overline{\pi}_h^k(s_h^k))  }  + \frac{\rho \sqrt{9SH^4\iota/2}}{N_h^k(s_h^k,\underline{\pi}_h^k(s_h^k))  }  \\
= &  (1-\rho) \sqrt{\frac{\mathbbm{V}_{{P}_h}{C}^{{\pi}^{k*},\underline{\pi}^k,\rho}_{h+1}(s_h^k,\overline{\pi}_h^k(s_h^k))\iota }{N_h^k(s_h^k, \overline{\pi}_h^k(s_h^k))  }} + \rho \sqrt{\frac{\mathbbm{V}_{{P}_h}{C}^{{\pi}^{k*},\underline{\pi}^k,\rho}_{h+1}(s_h^k,\underline{\pi}_h^k(s_h^k))\iota }{N_h^k(s_h^k, \underline{\pi}_h^k(s_h^k))  }} \\
& + \frac{\mathbbm{D}_{\widetilde{\pi}_h^k}P_h(\overline{V}^k_{h+1} - \underline{V}^k_{h+1})(s_h^k) }{2\sqrt{2}H} + \frac{2\sqrt{2}(1-\rho)H^2\iota}{N_h^k(s_h^k,\overline{\pi}_h^k(s_h^k))  } + \frac{2\sqrt{2}\rho H^2\iota}{N_h^k(s_h^k,\underline{\pi}_h^k(s_h^k))  } \\
&+ (1-\rho)\sqrt{\frac{1}{N_h^k(s_h^k, \overline{\pi}_h^k(s_h^k))  } }  +\rho\sqrt{\frac{1}{N_h^k(s_h^k, \underline{\pi}_h^k(s_h^k))  } } \\
& + \frac{(1-\rho) \sqrt{9SH^4\iota/2}}{N_h^k(s_h^k,\overline{\pi}_h^k(s_h^k))  }  + \frac{\rho \sqrt{9SH^4\iota/2}}{N_h^k(s_h^k,\underline{\pi}_h^k(s_h^k))  },
\end{split}    
\end{equation}
where the second inequality is due to AM-GM inequality.

\paragraph{Recursing on $h$} Plugging \eqref{eq:mbeepe}  and \eqref{eq:mbeeve} into \eqref{eq:regdecmboverh}and setting $c_1 = 1+ 1/H$ and $c_2 = (1+1/H)^3$ , we have 
\begin{equation} 
    \begin{split}
& \overline{V}_{h}^k(s_h^k) -  \underline{V}_h^{k}(s_h^k) \\
\le & (1+1/H)[ \mathbbm{D}_{\widetilde{\pi}_h^k}\hat{P}_h^k(\overline{V}_{h+1}^k - \underline{V}_{h+1}^k)](s_h^k) -(1+1/H) [\hat{P}_h^k(\overline{V}_{h+1}^k - \underline{V}_{h+1}^k)](s_h^k,a_h^k) \\\
& + (1/H+1/H^2){P}_h(\overline{V}_{h+1}^k - \underline{V}_{h+1}^k)(s_h^k,a_h^k) +\frac{(SH+SH^2)\iota}{N_h^k(s_h^k,a_h^k)}\\
& + c_1{P}_h(\overline{V}_{h+1}^k - \underline{V}_{h+1}^k)(s_h^k,a_h^k) -  c_2(\overline{V}_{h+1}^k - \underline{V}_{h+1}^k)(s_{h+1}^k) + c_2 (\overline{V}_{h+1}^k - \underline{V}_{h+1}^k)(s_{h+1}^k) \\
& + 2(1-\rho) \sqrt{\frac{2 \hat{r}_h^k(s_h^k,\overline{\pi}_h^k(s_h^k)) \iota}{N_h^k(s_h^k,\overline{\pi}_h^k(s_h^k))  }}  + \frac{2(1-\rho)(24H^2+7H+7)\iota}{3N_h^k(s_h^k,\overline{\pi}_h^k(s_h^k)) )} \\
&  + 2\rho\sqrt{\frac{2 \hat{r}_h^k(s_h^k,\underline{\pi}_h^k(s_h^k)) \iota}{N_h^k(s_h^k,\underline{\pi}_h^k(s_h^k))  }}  + \frac{2\rho(24H^2+7H+7)\iota}{3N_h^k(s_h^k,\underline{\pi}_h^k(s_h^k)) )} \\
& +  (1-\rho) \sqrt{\frac{8\mathbbm{V}_{{P}_h}{C}^{{\pi}^{k*},\underline{\pi}^k,\rho}_{h+1}(s_h^k,\overline{\pi}_h^k(s_h^k))\iota }{N_h^k(s_h^k, \overline{\pi}_h^k(s_h^k))  }} + \rho \sqrt{\frac{8\mathbbm{V}_{{P}_h}{C}^{{\pi}^{k*},\underline{\pi}^k,\rho}_{h+1}(s_h^k,\underline{\pi}_h^k(s_h^k))\iota }{N_h^k(s_h^k, \underline{\pi}_h^k(s_h^k))  }} \\
& + \frac{\mathbbm{D}_{\widetilde{\pi}_h^k}P_h(\overline{V}^k_{h+1} - \underline{V}^k_{h+1})(s_h^k) }{H} + \frac{8(1-\rho)H^2\iota}{N_h^k(s_h^k,\overline{\pi}_h^k(s_h^k))  } + \frac{8\rho H^2\iota}{N_h^k(s_h^k,\underline{\pi}_h^k(s_h^k))  } \\
&+ (1-\rho)\sqrt{\frac{8}{N_h^k(s_h^k, \overline{\pi}_h^k(s_h^k))  } }  +\rho\sqrt{\frac{8}{N_h^k(s_h^k, \underline{\pi}_h^k(s_h^k))  } }   + \frac{6(1-\rho) \sqrt{SH^4\iota}}{N_h^k(s_h^k,\overline{\pi}_h^k(s_h^k))  }  + \frac{6\rho \sqrt{SH^4\iota}}{N_h^k(s_h^k,\underline{\pi}_h^k(s_h^k))  } .
\end{split}
\end{equation}

We set $\Theta_h^k(s,a) = \sqrt{\frac{8\mathbbm{V}_{{P}_h}{C}^{{\pi}^{k*},\underline{\pi}^k,\rho}_{h+1}(s,a)\iota }{N_h^k(s,a)  }} + \sqrt{\frac{32}{N_h^k(s,a)  } } + \frac{46 \sqrt{SH^4\iota}}{N_h^k(s,a)  } $. Since $r_h^k(s,a) \le 1$, by organizing the items, we have that
\begin{equation} 
    \begin{split}
& \overline{V}_{h}^k(s_h^k) -  \underline{V}_h^{k}(s_h^k) \\
\le & (1+1/H)[ \mathbbm{D}_{\widetilde{\pi}_h^k}\hat{P}_h^k(\overline{V}_{h+1}^k - \underline{V}_{h+1}^k)](s_h^k) -(1+1/H) [\hat{P}_h^k(\overline{V}_{h+1}^k - \underline{V}_{h+1}^k)](s_h^k,a_h^k) \\
& + (1/H+1/H^2){P}_h(\overline{V}_{h+1}^k - \underline{V}_{h+1}^k)(s_h^k,a_h^k) +\frac{(SH+SH^2)\iota}{N_h^k(s_h^k,a_h^k)}\\
& + c_1{P}_h(\overline{V}_{h+1}^k - \underline{V}_{h+1}^k)(s_h^k,a_h^k) -  c_2(\overline{V}_{h+1}^k - \underline{V}_{h+1}^k)(s_{h+1}^k) + c_2 (\overline{V}_{h+1}^k - \underline{V}_{h+1}^k)(s_{h+1}^k) \\
& + \frac{\mathbbm{D}_{\widetilde{\pi}_h^k}P_h(\overline{V}^k_{h+1} - \underline{V}^k_{h+1})(s_h^k,\overline{\pi}_h^k(s_h^k)) }{H} + \mathbbm{D}_{\widetilde{\pi}_h^k}\Theta_h^k(s_h^k)\\
\le &(1+1/H)[ \mathbbm{D}_{\widetilde{\pi}_h^k}\hat{P}_h^k(\overline{V}_{h+1}^k - \underline{V}_{h+1}^k)](s_h^k) -(1+1/H) [\hat{P}_h^k(\overline{V}_{h+1}^k - \underline{V}_{h+1}^k)](s_h^k,a_h^k) \\
& + \frac{1}{H}[\mathbbm{D}_{\widetilde{\pi}_h^k}P_h(\overline{V}^k_{h+1} - \underline{V}^k_{h+1})(s_h^k)- {P}_h(\overline{V}_{h+1}^k - \underline{V}_{h+1}^k)(s_h^k,a_h^k) ] \\
& + (1+3/H +1/H^2){P}_h(\overline{V}_{h+1}^k - \underline{V}_{h+1}^k)(s_h^k,a_h^k) -  c_2(\overline{V}_{h+1}^k - \underline{V}_{h+1}^k)(s_{h+1}^k) \\
& + c_2 (\overline{V}_{h+1}^k - \underline{V}_{h+1}^k)(s_{h+1}^k)  +\frac{(SH+SH^2)\iota}{N_h^k(s_h^k,a_h^k)} + \mathbbm{D}_{\widetilde{\pi}_h^k}\Theta_h^k(s_h^k) \\
\le& (1+1/H)[ \mathbbm{D}_{\widetilde{\pi}_h^k}\hat{P}_h^k(\overline{V}_{h+1}^k - \underline{V}_{h+1}^k)](s_h^k) -(1+1/H) [\hat{P}_h^k(\overline{V}_{h+1}^k - \underline{V}_{h+1}^k)](s_h^k,a_h^k) \\
& + \frac{1}{H}[\mathbbm{D}_{\widetilde{\pi}_h^k}P_h(\overline{V}^k_{h+1} - \underline{V}^k_{h+1})(s_h^k)- {P}_h(\overline{V}_{h+1}^k - \underline{V}_{h+1}^k)(s_h^k,a_h^k) ] \\
& + c_2 {P}_h(\overline{V}_{h+1}^k - \underline{V}_{h+1}^k)(s_h^k,a_h^k) -  c_2(\overline{V}_{h+1}^k - \underline{V}_{h+1}^k)(s_{h+1}^k) \\
& + c_2 (\overline{V}_{h+1}^k - \underline{V}_{h+1}^k)(s_{h+1}^k)  +\frac{(SH+SH^2)\iota}{N_h^k(s_h^k,a_h^k)} + \mathbbm{D}_{\widetilde{\pi}_h^k}\Theta_h^k(s_h^k).
\end{split}
\end{equation}

By induction of \eqref{eq:regdecmboverh} on $h = 1, \cdots, H$ and $\overline{V}_{h+1}^k = \underline{V}_{h+1}^k =0$, we have that

\begin{equation}
    \begin{split}
        Regret(K) \le 21 \sum_{k=1}^K \sum_{h=1}^H (& \mathbbm{D}_{\widetilde{\pi}_h^k}\hat{P}_h^k(\overline{V}^k_{h+1} - \underline{V}^k_{h+1})(s_h^k)- \hat{P}_h^k(\overline{V}_{h+1}^k - \underline{V}_{h+1}^k)(s_h^k,a_h^k) \\
        & + \frac{1}{H}[\mathbbm{D}_{\widetilde{\pi}_h^k}P_h(\overline{V}^k_{h+1} - \underline{V}^k_{h+1})(s_h^k)- {P}_h(\overline{V}_{h+1}^k - \underline{V}_{h+1}^k)(s_h^k,a_h^k) ] \\
        & + {P}_h(\overline{V}_{h+1}^k - \underline{V}_{h+1}^k)(s_h^k,a_h^k) - (\overline{V}_{h+1}^k - \underline{V}_{h+1}^k)(s_{h+1}^k) \\
        &+\frac{(SH+SH^2)\iota}{N_h^k(s_h^k,a_h^k)}  + \mathbbm{D}_{\widetilde{\pi}_h^k}\Theta_h^k(s_h^k) ).
    \end{split}
\end{equation}

Here we use $(1+1/H)^3H < 21$.

\subsubsection{Proof of Lemma \ref{lem:M1}}
Recall that  $M_1 = \sum_{k=1}^K \sum_{h=1}^H [\mathbbm{D}_{\widetilde{\pi}_h^k}\hat{P}_h^k(\overline{V}^k_{h+1} - \underline{V}^k_{h+1})(s_h^k)- \hat{P}_h^k(\overline{V}_{h+1}^k - \underline{V}_{h+1}^k)(s_h^k,a_h^k)]$.

Since $\mathbbm{E}_{a_h^k \sim \mathbbm{D}_{\widetilde{\pi}_h^k}} [\hat{P}_h^k(\overline{V}_{h+1}^k - \underline{V}_{h+1}^k)(s_h^k,a_h^k) ] = \mathbbm{D}_{\widetilde{\pi}_h^k}\hat{P}_h^k(\overline{V}^k_{h+1} - \underline{V}^k_{h+1})(s_h^k) $, we have that $\mathbbm{D}_{\widetilde{\pi}_h^k}\hat{P}_h^k(\overline{V}^k_{h+1} - \underline{V}^k_{h+1})(s_h^k)- \hat{P}_h^k(\overline{V}_{h+1}^k - \underline{V}_{h+1}^k)(s_h^k,a_h^k)$  is a martingale
difference sequence. By the Azuma-Hoeffding inequality, with probability
$1-\delta$, we have 
\begin{equation}
   \left |\sum_{k=1}^K \sum_{h=1}^H [\mathbbm{D}_{\widetilde{\pi}_h^k}\hat{P}_h^k(\overline{V}^k_{h+1} - \underline{V}^k_{h+1})(s_h^k)- \hat{P}_h^k(\overline{V}_{h+1}^k - \underline{V}_{h+1}^k)(s_h^k,a_h^k)]\right | \le H\sqrt{2HK\iota}.
\end{equation}

\subsubsection{Proof of Lemma \ref{lem:M2}}
Recall that $M_2 =  \sum_{k=1}^K \sum_{h=1}^H \frac{1}{H}[\mathbbm{D}_{\widetilde{\pi}_h^k}P_h(\overline{V}^k_{h+1} - \underline{V}^k_{h+1})(s_h^k)- {P}_h(\overline{V}_{h+1}^k - \underline{V}_{h+1}^k)(s_h^k,a_h^k) ] $.

Since $\mathbbm{E}_{a_h^k \sim \mathbbm{D}_{\widetilde{\pi}_h^k}} [{P}_h(\overline{V}_{h+1}^k - \underline{V}_{h+1}^k)(s_h^k,a_h^k) ] = \mathbbm{D}_{\widetilde{\pi}_h^k}P_h(\overline{V}^k_{h+1} - \underline{V}^k_{h+1})(s_h^k) $, we have that $\mathbbm{D}_{\widetilde{\pi}_h^k}P_h(\overline{V}^k_{h+1} - \underline{V}^k_{h+1})(s_h^k)- {P}_h(\overline{V}_{h+1}^k - \underline{V}_{h+1}^k)(s_h^k,a_h^k)$  is a martingale
difference sequence. By the Azuma-Hoeffding inequality, with probability
$1-\delta$, we have 
\begin{equation}
  \left  |\sum_{k=1}^K \sum_{h=1}^H  [\mathbbm{D}_{\widetilde{\pi}_h^k}P_h(\overline{V}^k_{h+1} - \underline{V}^k_{h+1})(s_h^k)- {P}_h(\overline{V}_{h+1}^k - \underline{V}_{h+1}^k)(s_h^k,a_h^k)] \right| \le H\sqrt{2HK\iota}.
\end{equation}

\subsubsection{Proof of Lemma \ref{lem:M3}}
Recall that $M_3 = \sum_{k=1}^K \sum_{h=1}^H ({P}_h^k(\overline{V}_{h+1}^k - \underline{V}_{h+1}^k)(s_h^k,a_h^k) - (\overline{V}_{h+1}^k - \underline{V}_{h+1}^k)(s_{h+1}^k) ) $.

Let the one-hot vector $\hat{\mathbbm{1}}^k_h(\cdot| s_h^k,a_h^k)$ to satisfy that $\hat{\mathbbm{1}}^k_h(s_{h+1}^k| s_h^k,a_h^k) = 1$ and $\hat{\mathbbm{1}}^k_h(s| s_h^k,a_h^k) = 0$ for $s \neq s_{h+1}^k$. Thus, $[({P}_h^k- \hat{\mathbbm{1}}^k_h)(\overline{V}_{h+1}^k - \underline{V}_{h+1}^k)](s_h^k,a_h^k)$  is a martingale
difference sequence. By the Azuma-Hoeffding inequality, with probability
$1-\delta$, we have 
\begin{equation}
   \left |\sum_{k=1}^K \sum_{h=1}^H  [({P}_h^k- \hat{\mathbbm{1}}^k_h)(\overline{V}_{h+1}^k - \underline{V}_{h+1}^k)](s_h^k,a_h^k) \right| \le H\sqrt{2HK\iota}.
\end{equation}

\subsubsection{Proof of Lemma \ref{lem:M4}}

We bounded  $M_4 = \sum_{k=1}^K \sum_{h=1}^H [\frac{(SH+SH^2)\iota}{N_h^k(s_h^k,a_h^k)}+\mathbbm{D}_{\widetilde{\pi}_h^k}\Theta_h^k(s_h^k)] $ by separately bounding the four items.

\paragraph{Bound $\sum_{k=1}^K \sum_{h=1}^H \frac{(SH+SH^2)\iota}{N_h^k(s_h^k,a_h^k)} $} We regroup the summands in a different way.
\begin{equation}
    \begin{split}
        \sum_{k=1}^K \sum_{h=1}^H \frac{(SH+SH^2)\iota}{N_h^k(s_h^k,a_h^k) } = (SH+SH^2)\iota \sum_{h=1}^H \sum_{(s,a)\in\mathcal{S}\times\mathcal{A}} \sum_{n=1}^{N_h^K(s,a)}\frac{1}{n} \le (SH+SH^2)SAH\iota^2.
    \end{split}
\end{equation}

Recall that $\Theta_h^k(s,a) = \sqrt{\frac{8\mathbbm{V}_{{P}_h}{C}^{{\pi}^{k*},\underline{\pi}^k,\rho}_{h+1}(s,a)\iota }{N_h^k(s,a)  }} + \sqrt{\frac{32}{N_h^k(s,a)  } } + \frac{46 \sqrt{SH^4\iota}}{N_h^k(s,a)  } $.

\paragraph{Bound $\sum_{k=1}^K \sum_{h=1}^H [(1-\rho) \sqrt{\frac{ 32 \iota}{N_h^k(s_h^k,\overline{\pi}_h^k(s_h^k))  }} + \rho \sqrt{\frac{ 32 \iota}{N_h^k(s_h^k,\underline{\pi}_h^k(s_h^k))  }}] $}

We regroup the summands in a different way. For any policy $\pi$, we have 
\begin{equation}
    \begin{split}
        \sum_{k=1}^K \sum_{h=1}^H  \sqrt{\frac{ 32 \iota}{N_h^k(s_h^k,\pi(s_h^k))  }} = \sum_{h=1}^H \sum_{(s,a)\in\mathcal{S}\times\mathcal{A}} \sum_{n=1}^{N_h^K(s,a)} \sqrt{\frac{32 \iota}{n}} \le 8H\sqrt{SAK\iota}.
    \end{split}
\end{equation}

\paragraph{Bound $\sum_{k=1}^K \sum_{h=1}^H [(1-\rho) \frac{46SH^2\iota}{N_h^k(s_h^k,\overline{\pi}_h^k(s_h^k))  } + \rho\frac{46SH^2\iota}{N_h^k(s_h^k,\underline{\pi}_h^k(s_h^k))  }]  $}

We regroup the summands in a different way. For any policy $\pi$, we have 
\begin{equation}
    \begin{split}
        \sum_{k=1}^K \sum_{h=1}^H \frac{46\sqrt{SH^4\iota}}{N_h^k(s_h^k,\pi(s_h^k)) } = 46\sqrt{SH^4\iota} \sum_{h=1}^H \sum_{(s,a)\in\mathcal{S}\times\mathcal{A}} \sum_{n=1}^{N_h^K(s,a)}\frac{1}{n} \le 46S^{\frac{3}{2}}AH^3\iota^2.
    \end{split}
\end{equation}

\paragraph{Bound $\sum_{k=1}^K \sum_{h=1}^H  \left[(1-\rho) \sqrt{\frac{8\mathbbm{V}_{{P}_h}{C}^{{\pi}^{k*},\underline{\pi}^k,\rho}_{h+1}(s_h^k,\overline{\pi}_h^k(s_h^k))\iota }{N_h^k(s_h^k,\overline{\pi}_h^k(s_h^k))  }} + \rho\sqrt{\frac{8\mathbbm{V}_{{P}_h}{C}^{{\pi}^{k*},\underline{\pi}^k,\rho}_{h+1}(s_h^k,\underline{\pi}_h^k(s_h^k))\iota }{N_h^k(s_h^k,\underline{\pi}_h^k(s_h^k))  }} \right] $}

By Cauchy-Schwarz inequality,
\begin{equation}
    \begin{split}
& \sum_{k=1}^K \sum_{h=1}^H  \sqrt{\frac{\mathbbm{V}_{{P}_h}{C}^{{\pi}^{k*},\underline{\pi}^k,\rho}_{h+1}(s_h^k,\overline{\pi}_h^k(s_h^k))\iota }{N_h^k(s_h^k,\overline{\pi}_h^k(s_h^k))  }} \\
\le & \sqrt{  \sum_{k=1}^K \sum_{h=1}^H  \mathbbm{V}_{{P}_h}{C}^{{\pi}^{k*},\underline{\pi}^k,\rho}_{h+1}(s_h^k,\overline{\pi}_h^k(s_h^k)) \cdot \sum_{k=1}^K \sum_{h=1}^H  \frac{\iota}{N_h^k(s_h^k,\overline{\pi}_h^k(s_h^k))  }} \\
\le & \sqrt{ SAH\iota^2 \sum_{k=1}^K \sum_{h=1}^H  \mathbbm{V}_{{P}_h}{C}^{{\pi}^{k*},\underline{\pi}^k,\rho}_{h+1}(s_h^k,\overline{\pi}_h^k(s_h^k))  }.
    \end{split}
\end{equation}

Similarly, 
\begin{equation}
    \begin{split}
& \sum_{k=1}^K \sum_{h=1}^H  \sqrt{\frac{\mathbbm{V}_{{P}_h}{C}^{{\pi}^{k*},\underline{\pi}^k,\rho}_{h+1}(s_h^k,\underline{\pi}_h^k(s_h^k))\iota }{N_h^k(s_h^k,\underline{\pi}_h^k(s_h^k))  }} \\
\le & \sqrt{ SAH\iota^2 \sum_{k=1}^K \sum_{h=1}^H  \mathbbm{V}_{{P}_h}{C}^{{\pi}^{k*},\underline{\pi}^k,\rho}_{h+1}(s_h^k,\underline{\pi}_h^k(s_h^k))  }.
    \end{split}
\end{equation}

By $(1-\rho)a^2 + \rho b^2 \ge ((1-\rho)a + \rho b)^2$,
\begin{equation}
    \begin{split}
&(1-\rho) \sqrt{\sum_{k=1}^K \sum_{h=1}^H  \mathbbm{V}_{{P}_h}{C}^{{\pi}^{k*},\underline{\pi}^k,\rho}_{h+1}(s_h^k,\overline{\pi}_h^k(s_h^k)) } + \rho\sqrt{\sum_{k=1}^K \sum_{h=1}^H \mathbbm{V}_{{P}_h}{C}^{{\pi}^{k*},\underline{\pi}^k,\rho}_{h+1}(s_h^k,\underline{\pi}_h^k(s_h^k)) } \\
\le & \sqrt{\sum_{k=1}^K \sum_{h=1}^H [ (1-\rho)\mathbbm{V}_{{P}_h}{C}^{{\pi}^{k*},\underline{\pi}^k,\rho}_{h+1}(s_h^k,\overline{\pi}_h^k(s_h^k))+\rho\mathbbm{V}_{{P}_h}{C}^{{\pi}^{k*},\underline{\pi}^k,\rho}_{h+1}(s_h^k,\underline{\pi}_h^k(s_h^k)) ]  }.
    \end{split}
\end{equation}

Now we bound the total variance. Let $\mathbbm{D}_{\widetilde{\pi}_h^k}P_h(s'|s) = (1-\rho)P_h(s'|s,\overline{\pi}_h^k(s) +  \rho P_h(s'|s, \underline{\pi}_h^k(s) )$,
\begin{equation}
\begin{split}
        [\mathbbm{D}_{\widetilde{\pi}_h^k}P_h V_{h+1}] (s) = & \sum_{s'}[(1-\rho)P_h(s'|s,\overline{\pi}_h^k(s) )+ \rho P_h(s'|s, \underline{\pi}_h^k(s) )]V_{h+1}(s'),
\end{split}
\end{equation}
and
\begin{equation}
\begin{split}
        \mathbbm{V}_{[\mathbbm{D}_{\widetilde{\pi}_h^k}P_h]} V_{h+1} (s) = & \sum_{s'}[(1-\rho)P_h(s'|s,\overline{\pi}_h^k(s) )+ \rho P_h(s'|s, \underline{\pi}_h^k(s) )][V_{h+1}(s')]^2  \\
        & - [\sum_{s'}\left((1-\rho)P_h(s'|s,\overline{\pi}_h^k(s) )+ \rho P_h(s'|s, \underline{\pi}_h^k(s) )\right)V_{h+1}(s')]^2 .
\end{split}
\end{equation}
We have that
\begin{equation}
\begin{split}
& \mathbbm{V}_{[\mathbbm{D}_{\widetilde{\pi}_h^k}P_h]} {C}^{{\pi}^{k*},\underline{\pi}^k,\rho}_{h+1} (s_h^k)\\
    =&\sum_{s'}[(1-\rho)P_h(s'|s_h^k,\overline{\pi}_h^k(s_h^k) )+ \rho P_h(s'|s_h^k, \underline{\pi}_h^k(s_h^k) )][{C}^{{\pi}^{k*},\underline{\pi}^k,\rho}_{h+1}(s')]^2 \\
    & - [\sum_{s'}\left((1-\rho)P_h(s'|s_h^k,\overline{\pi}_h^k(s_h^k) )+ \rho P_h(s'|s_h^k, \underline{\pi}_h^k(s_h^k) )\right){C}^{{\pi}^{k*},\underline{\pi}^k,\rho}_{h+1}(s')]^2 \\
    \ge & (1-\rho)\mathbbm{V}_{{P}_h}{C}^{{\pi}^{k*},\underline{\pi}^k,\rho}_{h+1}(s_h^k,\overline{\pi}_h^k(s_h^k)) +\rho\mathbbm{V}_{{P}_h}{C}^{{\pi}^{k*},\underline{\pi}^k,\rho}_{h+1}(s_h^k,\underline{\pi}_h^k(s_h^k))\\
   &  + (1-\rho) [P_h{C}^{{\pi}^{k*},\underline{\pi}^k,\rho}_{h+1}(s_h^k,\overline{\pi}_h^k(s_h^k))]^2  + \rho P_h [ {C}^{{\pi}^{k*},\underline{\pi}^k,\rho}_{h+1}(s_h^k,\underline{\pi}_h^k(s_h^k))]^2 \\
    & - [\sum_{s'}(1-\rho)P_h(s'|s_h^k,\overline{\pi}_h^k(s_h^k) ){C}^{{\pi}^{k*},\underline{\pi}^k,\rho}_{h+1}(s') + \rho P_h(s'|s_h^k, \underline{\pi}_h^k(s_h^k) ){C}^{{\pi}^{k*},\underline{\pi}^k,\rho}_{h+1}(s')]^2 \\
    \ge& (1-\rho)\mathbbm{V}_{{P}_h}{C}^{{\pi}^{k*},\underline{\pi}^k,\rho}_{h+1}(s_h^k,\overline{\pi}_h^k(s_h^k)) +\rho\mathbbm{V}_{{P}_h}{C}^{{\pi}^{k*},\underline{\pi}^k,\rho}_{h+1}(s_h^k,\underline{\pi}_h^k(s_h^k)),
\end{split}
\end{equation}
where the last inequality is due to $(1-\rho)a^2 + \rho b^2 \ge ((1-\rho)a + \rho b)^2$.

With probability $1-2\delta$, we also have that
\begin{equation}
\begin{split}
  &   \sum_{k=1}^K \sum_{h=1}^H \mathbbm{V}_{[\mathbbm{D}_{\widetilde{\pi}_h^k}P_h]} {C}^{{\pi}^{k*},\underline{\pi}^k,\rho}_{h+1} (s_h^k) \\
  = & \sum_{k=1}^K \sum_{h=1}^H \left( [\mathbbm{D}_{\widetilde{\pi}_h^k}P_h({C}^{{\pi}^{k*},\underline{\pi}^k,\rho}_{h+1})^2](s_h^k)  - \left( [\mathbbm{D}_{\widetilde{\pi}_h^k}P_h{C}^{{\pi}^{k*},\underline{\pi}^k,\rho}_{h+1}](s_h^k)\right)^2    \right) \\
  = & \sum_{k=1}^K \sum_{h=1}^H \left( [\mathbbm{D}_{\widetilde{\pi}_h^k}P_h({C}^{{\pi}^{k*},\underline{\pi}^k,\rho}_{h+1})^2](s_h^k)  - \left( {C}^{{\pi}^{k*},\underline{\pi}^k,\rho}_{h+1}(s_{h+1}^k)\right)^2    \right) \\
 &   + \sum_{k=1}^K \sum_{h=1}^H \left( \left( {C}^{{\pi}^{k*},\underline{\pi}^k,\rho}_{h+1}(s_{h+1}^k)\right)^2   - \left( [\mathbbm{D}_{\widetilde{\pi}_h^k}P_h{C}^{{\pi}^{k*},\underline{\pi}^k,\rho}_{h+1}](s_h^k)\right)^2    \right) \\
\le & H^2\sqrt{2HK\iota}+ \sum_{k=1}^K \sum_{h=1}^H \left(({C}^{{\pi}^{k*},\underline{\pi}^k,\rho}_{h}(s_{h}^k))^2 - \left( [\mathbbm{D}_{\widetilde{\pi}_h^k}P_h{C}^{{\pi}^{k*},\underline{\pi}^k,\rho}_{h+1}](s_h^k)\right)^2 \right) -  \sum_{k=1}^K({C}^{{\pi}^{k*},\underline{\pi}^k,\rho}_{1}(s_{1}^k))^2 \\
\le & H^2\sqrt{2HK\iota}+ 2H \sum_{k=1}^K \sum_{h=1}^H |{C}^{{\pi}^{k*},\underline{\pi}^k,\rho}_{h}(s_{h}^k) - \mathbbm{D}_{\widetilde{\pi}_h^k}{P}_h{C}^{{\pi}^{k*},\underline{\pi}^k,\rho}_{h+1}(s_h^k) | \\
\le & H^2\sqrt{2HK\iota}+ 2H \sum_{k=1}^K \left({C}^{{\pi}^{k*},\underline{\pi}^k,\rho}_{1}(s_{1}^k) + \sum_{h=1}^H \left( {C}^{{\pi}^{k*},\underline{\pi}^k,\rho}_{h+1}(s_{h+1}^k) - \mathbbm{D}_{\widetilde{\pi}_h^k} {P}_h{C}^{{\pi}^{k*},\underline{\pi}^k,\rho}_{h+1}(s_h^k,a_h^k) \right) \right)\\
\le & H^2\sqrt{2HK\iota}+ 2H^2K + 2H^2\sqrt{2HK\iota} \\
\le& 3H^2K + 9H^3\iota/2, \\
\end{split}
\end{equation}
where the first inequality holds with probability $1-\delta$ by Azuma-Hoeffding inequality, the second inequality is due to the bound of V-values, the third inequality is due to Lemma~\ref{lem:Monot_mb} so that ${C}^{{\pi}^{k*},\underline{\pi}^k,\rho}_{h}(s_{h}^k) \ge \mathbbm{D}_{\widetilde{\pi}_h^k} D_h^{{\pi}^{k*},\underline{\pi}^k,\rho}(s_h^k) \ge \mathbbm{D}_{\widetilde{\pi}_h^k}{P}_h{C}^{{\pi}^{k*},\underline{\pi}^k,\rho}_{h+1}(s_h^k)$,  the fourth inequality holds with probability $1-\delta$ by Azuma-Hoeffding inequality, and the last inequality holds with $2ab \le a^2 + b^2$.

In summary, with probability at least $1-\delta$, we have
$\sum_{k=1}^K \sum_{h=1}^H  \mathbbm{V}_{{P}_h}{V}^{\overline{\pi}^k}_{h+1}(s_h^k,a_h^k) \le (H^2 K + H^3 \iota)$.

In summary, $\sum_{k=1}^K \sum_{h=1}^H \mathbbm{D}_{\widetilde{\pi}_h^k}\Theta_h^k(s_h^k) \le8\sqrt{SAH^2K\iota} +46S^{\frac{3}{2}}AH^3\iota^2 + \sqrt{24SAH^3K\iota^2+ 36SAH^5\iota^2} \le 8\sqrt{SAH^2K\iota} +46S^{\frac{3}{2}}AH^3\iota^2 +\sqrt{24SAH^3K}\iota +  6\sqrt{SAH^5}\iota $.

\section{Proof for model-free algorithm} \label{sec:prfmf}

In this section, we prove Theorem~\ref{thm:AR-UCBH}. Recall that we use 
$\overline{Q}_h^k$,$\overline{V}_h^k$,$\underline{Q}_h^k$,$\underline{V}_h^k$ and $N_h^k$ to denote the values of $\overline{Q}_h$,$\overline{V}_h$,$\underline{Q}_h$,$\underline{V}_h$ and $\max\{N_h,1\}$ at the beginning of the $k$-th episode.

\paragraph{Property of learning rate $\alpha_t$} We refer the readers to the setting of the learning rate $\alpha_t := \frac{H+1}{H+t}$ and the Lemma 4.1 in \cite{jin2018q}. For notational convenience, define $\alpha_t^0 := \prod_{j=1}^t (1-\alpha_t)$ and $\alpha_t^i := \alpha_i \prod_{j=i+1}^t (1-\alpha_t)$. 
Here, we introduce some useful properties of $\alpha_t^i$ which were proved in \cite{jin2018q}:\\
(1) $\sum_{i=1}^t \alpha_t^i= 1$ and $\alpha_t^0 = 0$ for $ t \ge 1$;\\
(2) $\sum_{i=1}^t \alpha_t^i= 0$ and $\alpha_t^0 =1$ for $t=0$;\\
(3) $ \frac{1}{\sqrt{t}} \le \sum_{i=1}^t \frac{\alpha_t^i}{\sqrt{t}} \le  \frac{2}{\sqrt{t}}$ for every $ t \ge 1$;\\
(4) $\sum_{i=1}^t (\alpha_t^i)^2 \le \frac{2H}{t}$ for every $ t \ge 1$;\\
(5) $\sum_{t=i}^\infty \alpha_t^i \le (1+\frac{1}{H})$ for every $ i \ge 1$.

\paragraph{Recursion on $Q$} As shown in \cite{jin2018q}, at any $(s, a, h, k) \in \mathcal{S} \times \mathcal{A} \times [H] \times [K]$, let $t = {N}_h^k(s,a)$ and suppose $(s,a)$ was previously taken by the agent at step $h$ of episodes $k_1, k_2, \dots, k_t < k$. By the update equations in Algorithm~\ref{alg:AR-UCBH} and the definition of $\alpha_t^i$, we have
\begin{equation}
    \begin{split}
     &   \overline{Q}_h^k(s,a) = \alpha_t^0 (H -h +1) + \sum_{i=1}^t \alpha_t^i \left(r_h^{k_i} + \overline{V}_{h+1}^{k_i}(s_{h+1}^{k_i})+b_i \right); \\
     & \underline{Q}_h^k(s,a) = \sum_{i=1}^t \alpha_t^i \left(r_h^{k_i} + \underline{V}_{h+1}^{k_i}(s_{h+1}^{k_i}) - b_i \right).
    \end{split}
\end{equation}

Thus, 
\begin{equation}
    \begin{split}
   (\overline{Q}_h^k- Q^*_h)(s,a)  =   &  \alpha_t^0(H -h +1) + \sum_{i=1}^t \alpha_t^i \left(r_h^{k_i} + \overline{V}_{h+1}^{k_i}(s_{h+1}^{k_i})+b_i \right) \\
    & - \left(\alpha_t^0 Q^*_h(s,a) + \sum_{i=1}^t \alpha_t^i \left(R_h(s,a) + P_h{V}_{h+1}^{*}(s,a) \right) \right) \\
     =  &  \alpha_t^0 (H -h +1 -  Q^*_h(s,a)) + \sum_{i=1}^t \alpha_t^i \left(  (\overline{V}_{h+1}^{k_i} -{V}_{h+1}^{*} ) (s_{h+1}^{k_i}) \right)  \\
     & + \sum_{i=1}^t \alpha_t^i \left( (r_h^{k_i} - R_h(s,a) ) + {V}_{h+1}^{*}(s_{h+1}^{k_i}) - P_h{V}_{h+1}^{*}(s,a)  +b_i \right),
    \end{split}
\end{equation}
and similarly
\begin{equation}
    \begin{split}
     (\underline{Q}_h^k -  Q^{\overline{\pi}^k}_h)(s,a)  = & \sum_{i=1}^t \alpha_t^i \left(r_h^{k_i} + \underline{V}_{h+1}^{k_i}(s_{h+1}^{k_i}) - b_i \right)  \\
    & - \left(\alpha_t^0 Q^{\overline{\pi}^k}_h(s,a) + \sum_{i=1}^t \alpha_t^i \left(R_h(s,a) + P_h{V}_{h+1}^{\overline{\pi}^k}(s,a) \right) \right)\\
     = &  - \alpha_t^0   Q^{\overline{\pi}^k}_h(s,a)+ \sum_{i=1}^t \alpha_t^i \left(   [P_h(\underline{V}_{h+1}^{k_i} - {V}_{h+1}^{\overline{\pi}^k})](s,a) \right) \\
     & + \sum_{i=1}^t \alpha_t^i \left( (r_h^{k_i} - R_h(s,a) ) + \underline{V}_{h+1}^{k_i}(s_{h+1}^{k_i}) - P_h\underline{V}_{h+1}^{k_i}(s,a) - b_i \right)  .
    \end{split}
\end{equation}

In addition, for any $k' \le k$, let $t' = {N}_h^{k'}(s,a)$. Thus, $(s,a)$ was previously taken by the agent at step $h$ of episodes $k_1, k_2, \dots, k_{t'} < k'$. We have
\begin{equation}
    \begin{split}
     (\underline{Q}_h^{k'} -  Q^{\overline{\pi}^k}_h)(s,a)  =  &  - \alpha_t^0   Q^{\overline{\pi}^k}_h(s,a)+ \sum_{i=1}^{t'} \alpha_{t'}^i \left(  [P_h(\underline{V}_{h+1}^{k_i} - {V}_{h+1}^{\overline{\pi}^k})](s,a) \right) \\
     &  + \sum_{i=1}^{t'} \alpha_{t'}^i \left( (r_h^{k_i} - R_h(s,a) ) + \underline{V}_{h+1}^{k_i}(s_{h+1}^{k_i}) - P_h\underline{V}_{h+1}^{k_i}(s,a) - b_i \right)  .
    \end{split}
\end{equation}

\paragraph{Confidence bounds}  By the Azuma-Hoeffding inequality, with probability $1-\delta$, we have that for all $s$, $a$, $h$ and $t \le K$,
\begin{equation}
   \left |\sum_{i=1}^t \alpha_t^i \left( (r_h^{k_i} - R_h(s,a) ) + \underline{V}_{h+1}^{k_i}(s_{h+1}^{k_i}) - P_h\underline{V}_{h+1}^{k_i}(s,a) \right) \right| \le H \sqrt{ \sum_{i=1}^t (\alpha_t^i)^2 \iota/2 } \le \sqrt{H^3\iota/t}.
\end{equation}
At the same time,  with probability $1-\delta$, we have that for all $s$, $a$, $h$ and $t \le K$,
\begin{equation}
   \left |\sum_{i=1}^t \alpha_t^i \left( (r_h^{k_i} - R_h(s,a) ) + {V}_{h+1}^{*}(s_{h+1}^{k_i}) - P_h{V}_{h+1}^{*}(s,a)\right) \right| \le \sqrt{H^3\iota/t}.
\end{equation}

In addition, we have $\sqrt{H^3\iota/t} \le \sum_{i=1}^t \alpha_t^i b_i \le 2\sqrt{H^3\iota/t}$.
{}

\paragraph{Monotonicity} Now we prove that $\overline{V}_h^k (s) \ge V_h^*(s) \ge V_h^{\overline{\pi}^k}(s) \ge \underline{V}_h^k (s)$ and $\overline{Q}_h^k (s,a) \ge Q_h^*(s,a) \ge Q_h^{\overline{\pi}^k}(s,a) \ge \underline{Q}_h^k (s,a)$ for all $(s, a, h, k) \in S \times A \times [H] \times[K]$.

At step $H+1$, we have  $\overline{V}_{H+1}^k (s) = V_{H+1}^*(s) = V_{H+1}^{\overline{\pi}^k}(s)= \underline{V}_{H+1}^k (s)  = 0$ and $\overline{Q}_{H+1}^k (s,a) = Q_{H+1}^*(s,a) = Q_{H+1}^{\overline{\pi}^k}(s,a) = \underline{Q}_{H+1}^k (s,a) = 0$ for all $(s, a, k) \in S \times A  \times[K]$.

Consider any step $h \in [H]$ in any episode $k \in [K]$, and suppose that the monotonicity is satisfied for all previous episodes as well as all steps $h' \ge h + 1$ in the current episode, which is 
\begin{equation}
\begin{split}
& \overline{V}_{h'}^{k'} (s) \ge V_{h'}^*(s) \ge V_{h'}^{\overline{\pi}^{k'}}(s) \ge \underline{V}_{h'}^{k'} (s) ~\forall (k',h',s) \in [k-1] \times [H+1] \times \mathcal{S}, \\
& \overline{Q}_{h'}^{k'} (s,a) \ge Q_{h'}^*(s,a) \ge Q_{h'}^{\overline{\pi}^{k'}}(s,a) \ge \underline{Q}_{h'}^{k'} (s,a) ~\forall (k',h',s,a) \in [k-1] \times [H+1] \times \mathcal{S} \times \mathcal{A}, \\
& \overline{V}_{h'}^k (s) \ge  V_{h'}^*(s) \ge V_{h'}^{\overline{\pi}^k}(s) \ge \underline{V}_{h'}^k (s) ~\forall h'\ge h+1 ~\text{and}~ s \in \mathcal{S}, 
 \\
& \overline{Q}_{h'}^k (s,a) \ge Q_{h'}^*(s,a) \ge Q_{h'}^{\overline{\pi}^k}(s,a) \ge \underline{Q}_{h'}^k (s,a) ~\forall h'\ge h+1 ~\text{and}~ (s,a) \in \mathcal{S}  \times \mathcal{A}. 
\end{split}
\end{equation}

We first show the monotonicity of $Q$ values.
We have 
\begin{equation}
\begin{split}
    (\overline{Q}_h^k- Q^*_h)(s,a) \ge  \alpha_t^0 (H - h +1 -  Q^*_h(s,a))  + \sum_{i=1}^t \alpha_t^i \left(  (\overline{V}_{h+1}^{k_i} -{V}_{h+1}^{*} ) (s_{h+1}^{k_i}) \right) \ge 0,
\end{split}
\end{equation}
and, by to the update rule of $\underline{V}$ values (line 13) in Algorithm~\ref{alg:AR-UCBH},
\begin{equation}
\begin{split}
   (\underline{Q}_h^k -  Q^{\overline{\pi}^k}_h)(s,a) \le  &   - \alpha_t^0 Q^{\overline{\pi}^k}_h(s,a) + \sum_{i=1}^t \alpha_t^i \left(  [P_h(\underline{V}_{h+1}^{k_i} - {V}_{h+1}^{\overline{\pi}^k})](s,a) \right)\\
   \le &  - \alpha_t^0 Q^{\overline{\pi}^k}_h(s,a) + \sum_{i=1}^t \alpha_t^i \left(  [P_h(\underline{V}_{h+1}^{k} - {V}_{h+1}^{\overline{\pi}^k})](s,a) \right) \le 0 .
\end{split}
\end{equation}
In addition, for any $k' \le k$,
\begin{equation}
\begin{split}
   (\underline{Q}_h^{k'} -  Q^{\overline{\pi}^k}_h)(s,a) \le  &   - \alpha_t^0 Q^{\overline{\pi}^k}_h(s,a) + \sum_{i=1}^{t'} \alpha_{t'}^i \left(  [P_h(\underline{V}_{h+1}^{k_i} - {V}_{h+1}^{\overline{\pi}^k})](s,a) \right)\\
   \le &  - \alpha_t^0 Q^{\overline{\pi}^k}_h(s,a) + \sum_{i=1}^{t'} \alpha_{t'}^i \left(  [P_h(\underline{V}_{h+1}^{k} - {V}_{h+1}^{\overline{\pi}^k})](s,a) \right)  \le 0 .
\end{split}
\end{equation}

Then, we show the monotonicity of $V$ values. We have that
\begin{equation}
    \begin{split}
& (1-\rho) \max_{a} \overline{Q}_h^k(s, a) + \rho \overline{Q}_h^k(s, \argmin_{a} \underline{Q}_h^k(s,a) ) \\
\ge &  (1-\rho) \max_{a} \overline{Q}_h^k(s, a) + \rho {Q}_h^*(s, \argmin_{a} \underline{Q}_h^k(s,a))\\
      \ge & (1-\rho)\overline{Q}_h^k(s, {\pi}_h^*(s)) + \rho \min_{a \in \mathcal{A}} {Q}_h^*(s, a) \\
      \ge & (1-\rho){Q}_h^*(s, {\pi}_h^*(s)) + \rho \min_{a \in \mathcal{A}} {Q}_h^*(s, a) = {V}_h^*(s).
    \end{split}
\end{equation}
By the update rule of $\overline{V}$ values (line 12) in Algorithm~\ref{alg:AR-UCBH}, \begin{equation} 
\overline{V}_{h}^k(s) = \min\{  \overline{V}_{h}^{k-1}(s),  (1-\rho) \max_{a} \overline{Q}_h^k(s, a) + \rho \overline{Q}_h^k(s, \argmin_{a} \underline{Q}_h^k(s,a) ) \} \ge {V}_h^*(s).
\end{equation}

Here, we need use the update rule of policy $\underline{\pi}$ (line 11-16) in Algorithm~\ref{alg:AR-UCBH}. 
Define ${\tau}(k,h,s) := \max\{k': k' < k ~\text{and}~ \underline{V}_{h}^{k'+1}(s)  = (1-\rho)\underline{Q}_h^{k'+1}(s,  \argmax_{a} \overline{Q}_h^{k'+1}(s, a)) + \rho \min_{a} \underline{Q}_h^{k'+1}(s, a)\}$, which denotes the last episode (before the beginning of the episode $k$), in which the $\overline{\pi}$ and $\underline{V}$ was updated at $(h,s)$. 
For notational simplicity, we use $\tau$ to denote ${\tau}(k,h,s)$ here. 
After the end of episode $\tau$ and before the beginning of the episode $k$, the agent policy $\overline{\pi}$ was not updated and $\underline{V}$ was not updated at $(h,s)$, i.e. $\underline{V}_{h}^k(s) = \underline{V}_{h}^{\tau+1}(s) = (1-\rho)\underline{Q}_h^{\tau+1}(s, \overline{\pi}_h^{\tau+1}(s)) + \rho \min_a \underline{Q}_h^{\tau+1}(s, a) $ and $\overline{\pi}_{h}^k(s) = \overline{\pi}_{h}^{\tau+1}(s) = \argmax_{a} \overline{Q}_h^{\tau+1}(s, a))$. Thus,
\begin{equation}
    \begin{split}
 \underline{V}_{h}^k(s) =  &  (1-\rho)\underline{Q}_h^{\tau+1}(s, \overline{\pi}_h^{\tau+1}(s)) + \rho \min_a \underline{Q}_h^{\tau+1}(s, a) \\
      \le & (1-\rho){Q}_h^{\overline{\pi}^k}(s, \overline{\pi}_h^{\tau+1}(s)) + \rho \min_a \underline{Q}_h^{\tau+1}(s, a)\\
      \le & (1-\rho){Q}_h^{\overline{\pi}^k}(s, \overline{\pi}_h^k(s)) + \rho \underline{Q}_h^{\tau+1}(s, \argmin_{a \in \mathcal{A}}{Q}_h^{\overline{\pi}^k}(s, a) ) \\
      \le & (1-\rho){Q}_h^{\overline{\pi}^k}(s, \overline{\pi}_h^k(s)) + \rho \min_{a \in \mathcal{A}} {Q}_h^{\overline{\pi}^k}(s, a) = {V}_h^{\overline{\pi}^k}(s).
    \end{split}
\end{equation}

By induction from $h = H+1$ to $1$ and $k = 1$ to $K$, we can conclude that $\overline{V}_h^k (s) \ge V_h^*(s) \ge V_h^{\overline{\pi}^k}(s) \ge \underline{V}_h^k (s)$ and $\overline{Q}_h^k (s,a) \ge Q_h^*(s,a) \ge Q_h^{\overline{\pi}^k}(s,a) \ge \underline{Q}_h^k (s,a)$ for all $(s, a, h, k) \in S \times A \times [H] \times[K]$.

\paragraph{Regret analysis} According to the monotonicity, the regret can be bounded by
\begin{equation}
    \begin{split}
        Regret(K):= & \sum_{k=1}^K ( V_{1}^*(s_1^k) -  {V}_1^{\overline{\pi}^k}(s_1^k))\le  \sum_{k=1}^K (\overline{V}_{1}^k(s_1^k) -  \underline{V}_1^{k}(s_1^k)) .
    \end{split}
\end{equation}
By the update rules in Algorithm~\ref{alg:AR-UCBH}, we have 
\begin{equation}
    \begin{split}
 &  \overline{V}_{h}^k(s_h^k) -  \underline{V}_h^{k}(s_h^k) \\
\le  & (1-\rho)  \overline{Q}_{h}^k(s_h^k,  \argmax_a \overline{Q}_{h}^k(s_h^k,a) ) + \rho \overline{Q}_{h}^k(s_h^k,  \argmin_a \underline{Q}_{h}^k(s_h^k,a) ) \\
& - (1-\rho)  \underline{Q}_{h}^k(s_h^k,  \argmax_a \overline{Q}_{h}^k(s_h^k,a) ) + \rho \underline{Q}_{h}^k(s_h^k,  \argmin_a \underline{Q}_{h}^k(s_h^k,a) ) \\
= & (1-\rho) [ \overline{Q}_{h}^k- \underline{Q}_{h}^k](s_h^k,  \overline{a}_{h}^k ) + \rho [ \overline{Q}_{h}^k- \underline{Q}_{h}^k](s_h^k, \underline{a}_{h}^k) \\
= &  [ \overline{Q}_{h}^k- \underline{Q}_{h}^k](s_h^k,  {a}_{h}^k ) + [ \mathbbm{D}_{\widetilde{\pi}_h^k}(\overline{Q}_{h}^k- \underline{Q}_{h}^k)](s_h^k) -[ \overline{Q}_{h}^k- \underline{Q}_{h}^k](s_h^k,  {a}_{h}^k ). 
\end{split}
\end{equation}
Set $n_h^k = N_h^k(s_h^k, a_h^k)$ and where $k_i(s_h^k,a_h^k)$ is the episode in which $(s_h^k,a_h^k)$ was taken at step $h$ for the $i$-th time. For notational simplicity, we set $\phi_h^k =  \overline{V}_{h}^k(s_h^k) -  \underline{V}_h^{k}(s_h^k)$ and $\xi_h^k = [ \mathbbm{D}_{\widetilde{\pi}_h^k}(\overline{Q}_{h}^k- \underline{Q}_{h}^k)](s_h^k) -[ \overline{Q}_{h}^k- \underline{Q}_{h}^k](s_h^k,  {a}_{h}^k ) $. According to the update rules,
\begin{equation}
    \begin{split}
\phi_h^k = &  \overline{V}_{h}^k(s_h^k) -  \underline{V}_h^{k}(s_h^k) \\
\le & \alpha_{n_h^k}^0 (H -h +1) + \sum_{i=1}^{n_h^k} \alpha_{n_h^k}^i \left(\overline{V}_{h+1}^{k_i(s_h^k,a_h^k)}(s_{h+1}^{k_i(s_h^k,a_h^k)})- \underline{V}_{h+1}^{k_i(s_h^k,a_h^k)}(s_{h+1}^{k_i(s_h^k,a_h^k)}) + 2 b_i \right) \\
& + [ \mathbbm{D}_{\widetilde{\pi}_h^k}(\overline{Q}_{h}^k- \underline{Q}_{h}^k)](s_h^k) -[ \overline{Q}_{h}^k- \underline{Q}_{h}^k](s_h^k,  {a}_{h}^k ) \\
= & \alpha_{n_h^k}^0 (H -h +1) + \sum_{i=1}^{n_h^k} \alpha_{n_h^k}^i (\phi_{h+1}^{k_i(s_h^k,a_h^k)} + 2 b_i ) + \xi_h^k \\
\le & \alpha_{n_h^k}^0 (H -h +1) + \sum_{i=1}^{n_h^k} \alpha_{n_h^k}^i \phi_{h+1}^{k_i(s_h^k,a_h^k)}  + \xi_h^k + 4\sqrt{H^3 \iota / n_h^k}.
\end{split}
\end{equation}

We add $\overline{V}_{h}^k(s_h^k) -  \underline{V}_h^{k}(s_h^k)$ over $k$ and regroup the summands in a different way. Note that for any episode $k$, the term $\sum_{i=1}^{n_h^k} \alpha_{n_h^k}^i \phi_{h+1}^{k_i(s_h^k,a_h^k)}$ takes all the prior episodes $k_i < k$ where $(s_h^k, a_h^k)$ was taken into account. In other words, for any episode $k'$, the term $\phi_{h+1}^{k'}$ appears in the summands at all posterior episodes $k > k'$ where $(s_h^{k'}, a_h^{k'})$ was taken. The first time it appears we have ${n}_{h}^{k} = n_{h}^{k'}+1$, and the second time it appears we have ${n}_{h}^{k} = n_{h}^{k'}+2$, and so on. Thus, we have 
\begin{equation}
    \begin{split}
 &   \sum_{k=1}^K (\overline{V}_{h}^k(s_h^k) -  \underline{V}_h^{k}(s_h^k)) \\
\le  &  \sum_{k=1}^K \alpha_{n_h^k}^0 (H -h +1) +  \sum_{k=1}^K \sum_{i=1}^{n_h^k} \alpha_{n_h^k}^i \phi_{h+1}^{k_i(s_h^k,a_h^k)}  +  \sum_{k=1}^K \xi_h^k +  \sum_{k=1}^K 4\sqrt{H^3 \iota / n_h^k} \\
= &  \sum_{k=1}^K \alpha_{n_h^k}^0 (H -h +1) +  \sum_{k'=1}^K \phi_{h+1}^{k'} \sum_{t=n_h^{k'}+1}^{n_h^K} \alpha_{t}^{n_h^{k'}}   +  \sum_{k=1}^K \xi_h^k +  \sum_{k=1}^K 4\sqrt{H^3 \iota / n_h^k} \\
\le &  \sum_{k=1}^K \alpha_{n_h^k}^0 (H -h +1) +  (1+1/H)\sum_{k=1}^K \phi_{h+1}^{k}  +  \sum_{k=1}^K \xi_h^k +  \sum_{k=1}^K 4\sqrt{H^3 \iota / n_h^k}
\end{split}
\end{equation}
where the final inequality uses the property $\sum_{t=i}^\infty \alpha_t^i \le (1+\frac{1}{H})$ for every $ i \ge 1$.

Taking the induction from $h = 1$ to $H$, we have
\begin{equation}
    \begin{split}
 &   \sum_{k=1}^K (\overline{V}_{1}^k(s_1^k) -  \underline{V}_1^{k}(s_1^k)) \\
\le & 3 \sum_{h=1}^H \sum_{k=1}^K \alpha_{n_h^k}^0 (H -h +1)   +  3 \sum_{h=1}^H \sum_{k=1}^K \xi_h^k +  \sum_{h=1}^H \sum_{k=1}^K 12 \sqrt{H^3 \iota / n_h^k}
\end{split}
\end{equation}
where we use the fact that $(1+1/H)^H < 3$ and $\phi_{H+1}^{k} = 0$ for all $k$.

We bound the three items separately.

(1) We have $\sum_{h=1}^H \sum_{k=1}^K \alpha_{n_h^k}^0 (H -h +1) =  \sum_{h=1}^H \sum_{k=1}^K \mathbbm{1}[n_h^k = 0] (H -h +1) \le SAH^2$.

(2) Similar to Lemma \ref{lem:M1}, by the Azuma-Hoeffding inequality, with probability
$1-\delta$, we have $\sum_{h=1}^H \sum_{k=1}^K \xi_h^k \le H\sqrt{2HK\iota}$.

(3) We have $\sum_{h=1}^H \sum_{k=1}^K 12 \sqrt{H^3 \iota / n_h^k} = \sum_{h=1}^H \sum_{(s,a)} \sum_{n=1}^{N_h^K(s,a)} \sqrt{H^3 \iota / n} \le H \sqrt{2H^3SAK\iota}$.

In summary,  $$Regret(K) = \sum_{k=1}^K ( V_{1}^*(s_1^k) -  {V}_1^{\overline{\pi}^k}(s_1^k)) \le \mathcal{O}(\sqrt{SAH^5K\iota} + SAH^2)$$  and 
\begin{equation}
\begin{split}
V_{1}^*(s_1) -  {V}_1^{{\pi}^{out}}(s_1) \le & \overline{V}_{1}^{K+1}(s_1) -  \underline{V}_1^{K+1}(s_1)  \\
=& \min_{k \in [K+1]} (\overline{V}_{1}^{k}(s_1^k) -  \underline{V}_1^{k}(s_1^k))\\ 
\le& O\left(\frac{\sqrt{SAH^5\iota}}{K} + \frac{SAH^2}{K}\right).
\end{split}
\end{equation}
\end{document}